\documentclass{article}

    \PassOptionsToPackage{numbers, compress}{natbib}

 \usepackage[main, final]{neurips_2026}

\usepackage[utf8]{inputenc} 
\usepackage[T1]{fontenc}    
\usepackage{hyperref}       
\usepackage{url}            
\usepackage{booktabs}       
\usepackage{amsfonts}       
\usepackage{nicefrac}       
\usepackage{microtype}      
\usepackage{xcolor}         
\usepackage{pifont}

\usepackage{makecell}

\usepackage{multicol}
\usepackage{subcaption}
\usepackage{mathrsfs}
\usepackage{amsfonts}
\usepackage{colortbl}               
\usepackage{amsfonts}       
\usepackage{nicefrac}       
\usepackage{microtype}  
\usepackage{marvosym}
\usepackage{array} 
\usepackage{graphicx}
\usepackage{amsmath}
\usepackage{amssymb}
\usepackage{multirow}
\setcitestyle{numbers,square}
\usepackage{algorithm}  
\usepackage{algorithmicx} 
\usepackage{algpseudocode}
\usepackage{caption}
\usepackage{subfloat}
\usepackage{tabularx} 
\usepackage{ragged2e} 
\usepackage{makecell}
\usepackage{amsfonts}
\usepackage{xspace}
\usepackage{anyfontsize}
\usepackage{nicefrac}  
\usepackage[accsupp]{axessibility} 
\usepackage{float}
\usepackage{cleveref}
\usepackage{wrapfig}

\title{
\raisebox{-0.3\height}{\includegraphics[height=1.2cm]{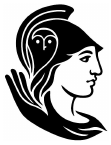}}\hspace{0.3em}METIS: Multi-Source Egocentric Training for Integrated Dexterous Vision-Language-Action Model
}

\author{Yankai Fu\textsuperscript{1,2}$^{*}$, 
        Ning Chen\textsuperscript{1,2}$^{*}$, 
        Junkai Zhao\textsuperscript{2}$^{*\dagger}$, 
        Shaozhe Shan\textsuperscript{1}, \\
        \textbf{Guocai Yao\textsuperscript{2}, 
        Pengwei Wang\textsuperscript{2}, 
        Zhongyuan Wang\textsuperscript{2}, 
        Shanghang Zhang\textsuperscript{1,2}\textsuperscript{\Letter}} \vspace{0.3em} \\
    \textsuperscript{1}State Key Laboratory of Multimedia Information Processing, School of Computer Science, \\
    Peking University; \textsuperscript{2}Beijing Academy of Artificial Intelligence \\
    $^{*}$Equal contribution, $^{\dagger}$Project leader, \textsuperscript{\Letter}Corresponding author
    \vspace{0.3em}\\
    {\textbf{Project Webpage:} \href{https://aureleopku.github.io/METIS}{https://aureleopku.github.io/METIS}
    }
}

\begin{document}

\maketitle

\begin{center}
    \captionsetup{type=figure}
    \includegraphics[width=\textwidth]{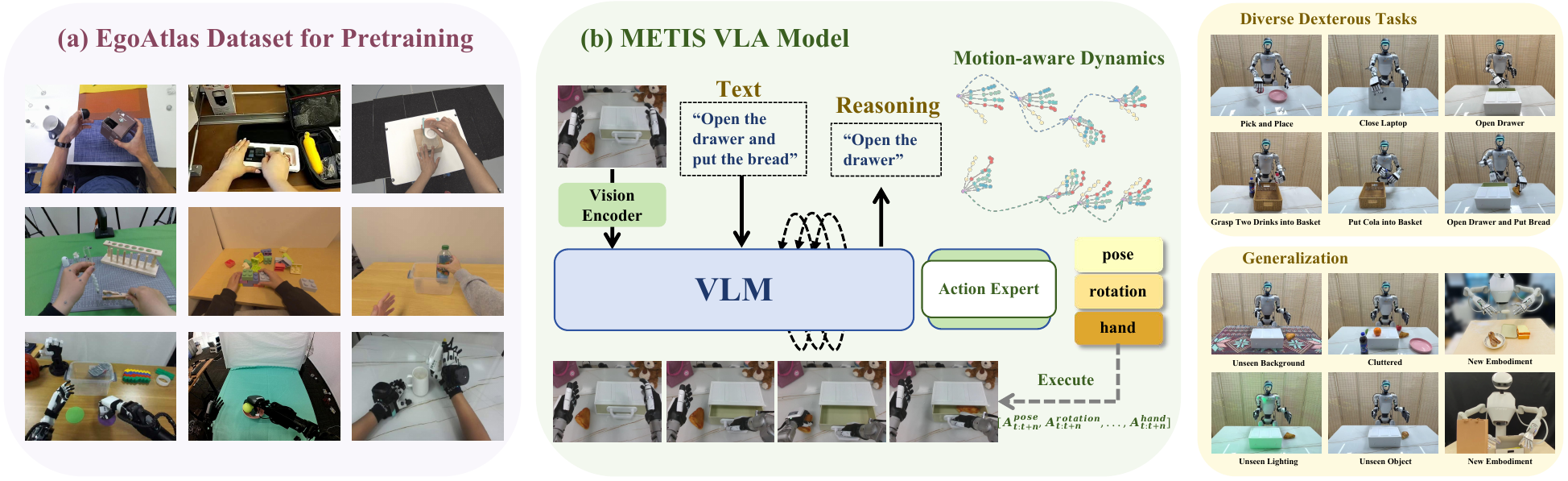}
    \captionof{figure}{METIS is trained on a multi-source egocentric manipulation dataset EgoAtlas. It leverages motion-aware dynamics to extract manipulation-relevant dexterous features, and integrates reasoning and acting within a unified framework. METIS achieves strong performance across diverse dexterous manipulation tasks and exhibits remarkable generalization capability.}
\end{center}

\begin{abstract}
  Building a generalist robot that can perceive, reason, and act across diverse tasks remains an open challenge, especially for dexterous manipulation. A major bottleneck lies in the scarcity of large-scale, action-annotated data for dexterous skills, as teleoperation is difficult and costly. Human data, with its vast scale and diverse manipulation behaviors, provides rich priors for learning robotic actions. While prior works have explored leveraging human demonstrations, they are often constrained by limited scenarios and a large visual gap between human and robots. To eliminate these limitations, we propose METIS, a vision-language-action (VLA) model for dexterous manipulation pretrained on multi-source egocentric datasets. We first construct EgoAtlas, which integrates large-scale human and robotic data from multiple sources, all unified under a consistent action space. We further extract motion-aware dynamics, a compact and discretized motion representation, which provides efficient and expressive supervision for VLA training. Built upon them, METIS integrates reasoning and acting into a unified framework, enabling effective deployment to downstream dexterous manipulation tasks. Our method demonstrates exceptional dexterous manipulation capabilities, achieving highest average success rate in eight real-world tasks. Experimental results also highlight the superior generalization and robustness to out-of-distribution scenarios. These findings emphasize METIS as a promising step toward a generalist model for dexterous manipulation.
\end{abstract}

\section{Introduction}
Recent advances in vision-language-action (VLA) models have achieved remarkable progress toward general-purpose embodied intelligence \cite{team2024octo, kim24openvla, intelligence2025pi_, zitkovich2023rt}. However, such models heavily rely on scarce and expensive real-world robot data, often collected via human teleoperation. Despite extensive efforts to build large-scale robotic datasets \cite{o2024open, khazatsky2024droid, bu2025agibot, wu2024robomind}, the effective scale of robot interaction data remains orders of magnitude lower than those used for LLM training \cite{zitkovich2023rt}, severely limiting the scalability and generalization of current VLAs. This issue is even more pronounced in dexterous manipulation, where acquiring high-quality demonstrations is extremely costly and complex. As a result, most existing VLA research focuses on simple gripper-based tasks \cite{huang2023embodied, li2023vision, bu2025univla, liu2025hybridvla}, leaving dexterous manipulation largely unexplored. 

In contrast to the scarcity of robotic datasets, human data are vastly abundant and rich in semantic information, providing valuable behavioral priors. Previous works have explored learning task-relevant representations from human videos, such as affordance \cite{bahl2023affordances, kuang2024ram}, latent action \cite{ye2024latent, bu2025univla}, and keypoint flow \cite{zhang2025actron3d, yuan2024general}. Recent progress such as EgoVLA \cite{yang2025egovla}, Being-H0 \cite{luo2025being}, and H-RDT \cite{bi2025h} leverage large-scale human datasets to pretrain VLA, providing a new paradigm for scaling robotic learning. However, human videos are often confined to specific household or workspace scenes (e.g., tabletop or kitchen), exhibiting uneven scene coverage and strong contextual biases. Moreover, there exists a substantial gap between human data and robot data, in both visual appearance and action space. Another line of research explores co-training strategies that jointly utilize self-collected human and robotic data. For example, HAT \cite{qiu2025humanoid} learns a human policy and demonstrates that incorporating human data can significantly enhance the generalization and robustness of robotic policies, while MotionTrans \cite{yuan2025motiontrans} aligns human motions to robot-specific embodiments to achieve zero-shot skill transfer. These approaches collect manipulation-related human data but do not fully exploit the vast amount of human data available on the internet.

In this work, we investigate these problems by introducing\textbf{ METIS, a vision-language-action (VLA) model} for dexterous manipulation pretrained on multi-source egocentric datasets. We first construct \textbf{EgoAtlas}, a multi-source egocentric dataset that covers large-scale internet human data, robot data, and enhanced human data collected through a wearable system. EgoAtlas spans four major categories and eight sources, all aligned under a unified action space. We further propose motion-aware dynamics, a compact and discretized representation designed for dexterous manipulation. It captures both visual and motion dynamics, providing efficient and expressive supervision for training VLA. Built upon them, METIS is pretrained on EgoAtlas, unifying reasoning and acting within a single framework. This design enables efficient fine-tuning and deployment on downstream dexterous tasks.

To comprehensively evaluate METIS, we conduct extensive real-world experiments on dexterous tasks. The comparative results show that our model achieves superior performance and efficiency, attaining the highest average success rate among all baseline VLAs. Besides, METIS also exhibits strong generalization to out-of-distribution scenarios, including unseen background, unseen object, unseen lighting condition, cluttered environment, and can be transferred to higher-DoF embodiments. Finally, ablation studies verify the contributions of both the multi-source egocentric dataset and the motion-aware dynamics to the overall system, highlighting the potential of learning robotic motion priors from human data.
In summary, our contributions are as follows:

\begin{itemize}
    \item We construct a multi-source egocentric manipulation dataset EgoAtlas, which integrates diverse human and robotic data sources under a unified state-action space. 
    \item We present METIS, a VLA model for dexterous manipulation pretrained on EgoAtlas, which extracts motion-aware dynamics from dexterous hand motion and integrates reasoning and acting within a unified framework.
    \item We demonstrate the effectiveness and generalization of our method through a range of real-world experiments.
\end{itemize}

\section{Related Work}
\label{sec:related_work}

\subsection{Dexterous Manipulation}
\label{subsec:dexterous_manipulation}

Dexterous manipulation is a key challenge in robotics, focusing on achieving human-like fine-grained manipulation skills. Traditional approaches are based on optimization and control algorithms \cite{kumar2014real, wang2022dexgraspnet, li2025learning, yang2025multi}, typically assuming access to known dynamics and object models, and focusing on trajectory planning under physical constraints. While achieving strong performance in specific scenarios, these methods often struggle to generalize across diverse real-world settings. Recent learning-based approaches, including reinforcement learning (RL) and imitation learning (IL), have made remarkable progress across various tasks such as grasping \cite{xu2023unidexgrasp, wan2023unidexgrasp++}, in-hand manipulation \cite{qi2023hand, pitz2023dextrous}, and tool use \cite{shaw2024bimanual, yin2025dexteritygen}. RL methods learn precise and dexterous skills through large-scale training in simulation, but often suffer from a significant sim-to-real gap. In contrast, IL methods leverage expert demonstrations to achieve robust performance in real-world dexterous manipulation tasks \cite{liconti2024leveraging, gbagbe2024bi, kim2024goal, fu2025cordvip}, but typically rely on expensive human teleoperation data. In this work, we address this data bottleneck by pretraining our model on large-scale multi-source egocentric data that combines human and robotic demonstration to learn motion priors efficiently.


\subsection{Learning Dexterity from Human Data}
\label{subsec:learning dexterity from human}
Human data offer valuable priors for dexterous manipulation, featuring abundant samples, fine-grained hand motion, and rich semantic cues. There has been works on learning task-relevant representations from human videos, such as affordance \cite{bahl2023affordances, kuang2024ram}, latent action \cite{ye2024latent, bu2025univla}, and keypoint flow \cite{zhang2025actron3d, yuan2024general}. Ego-Only \cite{wang2023ego} utilizes the MAE framework \cite{he2022masked} to extract actionable information from egocentric videos, while LAPA \cite{ye2024latent} and UniVLA \cite{bu2025univla} apply VQ-VAE \cite{van2017neural} to extract latent action from large-scale unlabeled human data. However, learning directly from human videos introduces redundant, manipulation-irrelevant content and underemphasizes the critical role of hand motion. To mitigate this gap, HAT \cite{qiu2025humanoid} and MotionTrans \cite{yuan2025motiontrans} adopt a co-training strategy on both human and robotic data with explicit motion information, improving the robustness of the policy and enabling zero-shot skill transfer.
In this work, we learn action priors from human data through joint modeling of visual and motion dynamics to support VLA learning.

\subsection{Vision-Language-Action Model}
\label{subsec:vision-language-action model}
Vision-Language-Action (VLA) models have achieved unprecedented progress in recent years \cite{team2024octo, kim24openvla, intelligence2025pi_, qu2025spatialvla}, driven by advances in Vision-Language Models (VLMs) \cite{team2025robobrain, team2025gemini, yang2025magma} and the availability of large-scale robotic datasets \cite{o2024open, khazatsky2024droid, bu2025agibot, wu2024robomind, hou2025robomind}. By processing multimodal inputs—such as visual observations and language instructions—these models enable robots to autonomously perform a wide range of tasks. Representative VLA models, such as RT-2 \cite{zitkovich2023rt} and OpenVLA \cite{kim24openvla} generate action sequences autoregressively. In contrast, $\pi_0$ \cite{black2024pi_0} and DexVLA \cite{wen2025dexvla} employ a diffusion-based policy, fitting continuous action distributions via iterative denoising. However, these approaches primarily focus on gripper-based manipulation, overlooking the rich motion and interaction dynamics inherent in dexterous tasks. Recently, several studies have extended VLA frameworks to dexterous manipulation. For example, GR00T N1 \cite{bjorck2025gr00t} learns latent representations from hybrid data to train a humanoid manipulation policy, while EgoVLA \cite{yang2025egovla} and Being-H0 \cite{luo2025being} leverage large-scale human demonstrations for VLA pretraining, enabling the acquisition of motion priors. Despite their impressive results, the scene homogeneity of human videos leads to bias and a large visual gap from robotic observations. To overcome this issue, we leverage multi-source egocentric data, introduce an enhanced human dataset, and develop an integrated VLA model for unified reasoning and acting.

\section{EgoAtlas Dataset}

\subsection{Wearable System for Enhanced Human Data}
\label{Wearable System for Enhanced Human Data}
Traditional hand motion capture systems rely on multi-camera or VR setups, which are limited by viewpoint dependency, occlusion, and constrained capture space. To overcome these issues, we develop a wearable glove–tracker system for portable, high-fidelity motion capture.


\noindent\textbf{Wearable Hand Motion Collection System.} 
We use Manus Quantum Metagloves to record precise 3D hand keypoints (25 per hand). A VIVE Tracker mounted on each glove provides the 6-DoF wrist pose for global localization. To capture egocentric visual observations, a head-mounted camera is used to 
record the first-person view during manipulation, providing visual information closely aligned with the operator’s perspective, while another VIVE Tracker is attached to the headset to perform extrinsic calibration. 
This calibration aligns the 
\begin{wrapfigure}{r}{0.44\linewidth}
    \centering
    \includegraphics[width=\linewidth]{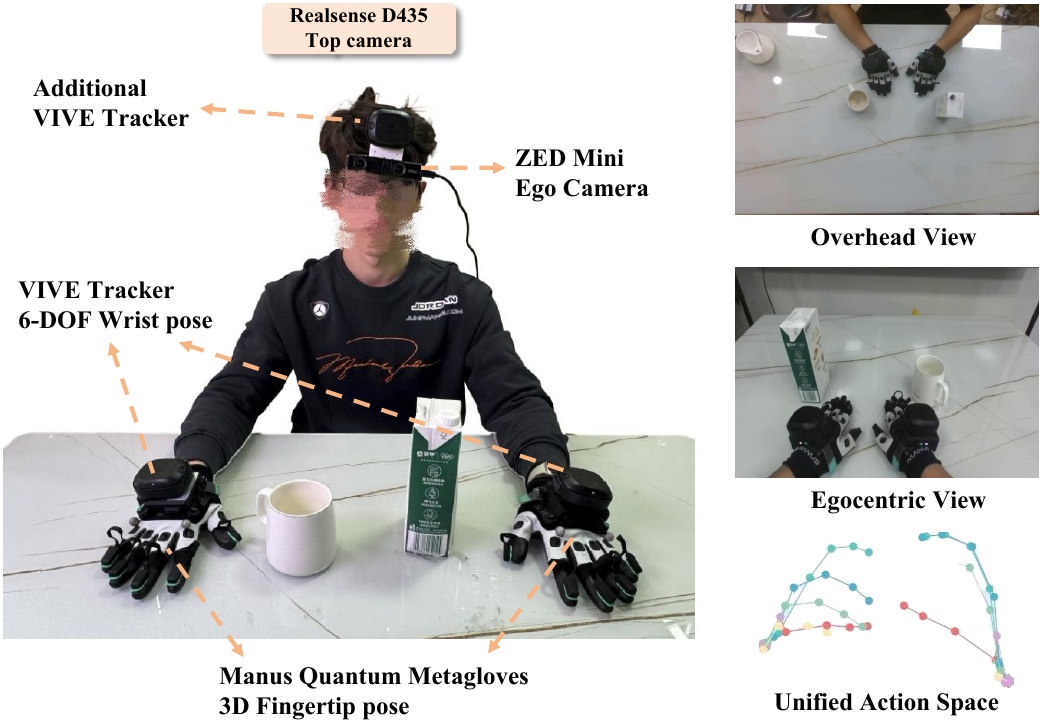}
    \caption{\textbf{Wearable Hand Motion Collection System.}}
    \label{fig:collection_system}
    \vspace{-2em}
\end{wrapfigure}motion-capture and ego-camera coordinate frames, enabling the entire system to operate in the wild and ensuring scene diversity. We also provide a top-down third-person camera view which is optional and is not used for calibration, training, or inference in this work; it records RGB videos only for future research purposes.
The entire system operates at 20~Hz, balancing motion fidelity with reliable multi-sensor synchronization.

\noindent\textbf{Subtask-Level Annotation.} We performed subtask-level annotation to enrich the semantic structure of the dataset. 
Each trajectory is paired with a task-level language instruction, along with a fine-grained segmentation into multiple subtasks. This annotation design enables the VLA to focus on hand motion and supports hierarchical reasoning over long-horizon manipulation tasks.
More details can be found in Appendix \ref{Appendix: Subtask-Level Annotation}.

\subsection{Data Statistics and Processing}

Built on \Cref{Wearable System for Enhanced Human Data}, we introduce EgoAtlas, a large-scale multi-source egocentric dataset for VLA training. 
EgoAtlas integrates data from 8 heterogeneous sources spanning human and robot domains, including 10K self-collected high-fidelity trajectories using wearable system.  The final dataset contains 343K trajectories and 89.72M image–action pairs,  as summarized in \Cref{tab:Statistics of EgoAtlas}. 
We apply weighted sampling to balance source distributions. 
More details are provided in Appendix \ref{Appendix: Data Sources and Statistics}.

Different embodiments operate in distinct state and action space.
To bridge these heterogeneous sources, we construct a unified state–action representation that abstracts away embodiment-specific differences while preserving manipulation-relevant motion. The state–action space is composed of two components: wrist pose and fingertip pose. The wrist pose is represented as an 18-dimensional vector consisting of a 3D position and a 6D continuous rotation per hand in the egocentric camera frame. The hand configuration is represented by a 30-dimensional fingertip pose, defined by the 3D positions of individual fingertips in the wrist coordinate frame.
We apply robot-specific FK/IK (Forward Kinematics and Inverse Kinematics) and human motion retargeting algorithms to project heterogeneous motion representations into a shared state–action space, while aligning all data within a consistent coordinate system. More details can be found in Appendix \ref{Appendix: Coordinate System Definition}, \ref{Appendix: Unifying Action Space} and \ref{Appendix: Hand IK Retargetting}.

\section{Method}


\subsection{Problem Formulation}
\label{Problem Formulation}
Our goal is to learn a VLA model for dexterous manipulation from multi-source egocentric data. Formally, given a collection of human and robot trajectories $D = D_{robot} \cup D_{human}$ from the EgoAtlas dataset, each trajectory can be represented as a sequence of observation–action pairs $\tau=\{(o_t,a_t)\}_{t=1}^{T}$, where observation $o_t = \{I_t, S_t\}$ consists of the egocentric image input and the proprioceptive state. The training objective is to learn a policy $\pi_{\phi}(a_t|o_t, l)$ that predicts actions conditioned on egocentric observations $o_t$ and language instruction $l$.

To bridge the embodiment gap, we construct a unified state–action space which includes an 18D wrist pose $P^w_t$ (3D position and 6D rotation vector per hand) and a 30D finger pose $P^f_t$ capturing the 3D position of individual fingertips. Here we describe the design of the three components in detail.



\begin{itemize}
    \item Image Observation $I_t$: We use egocentric image from a first-person camera, which provides a viewpoint that focuses on fine-grained interaction details.
    \item Wrist Pose $P^w_t$: We represent the wrist pose as the 3D position and the 6D rotation vector following \citet{zhou2019continuity}. They are relative to the camera coordinate frame.
    \item Finger Pose $P^f_t$: We represent the finger pose using 3D fingertip positions, defined in the wrist coordinate frame.
\end{itemize}

\begin{figure*}
    \centering
    \includegraphics[width=\linewidth]{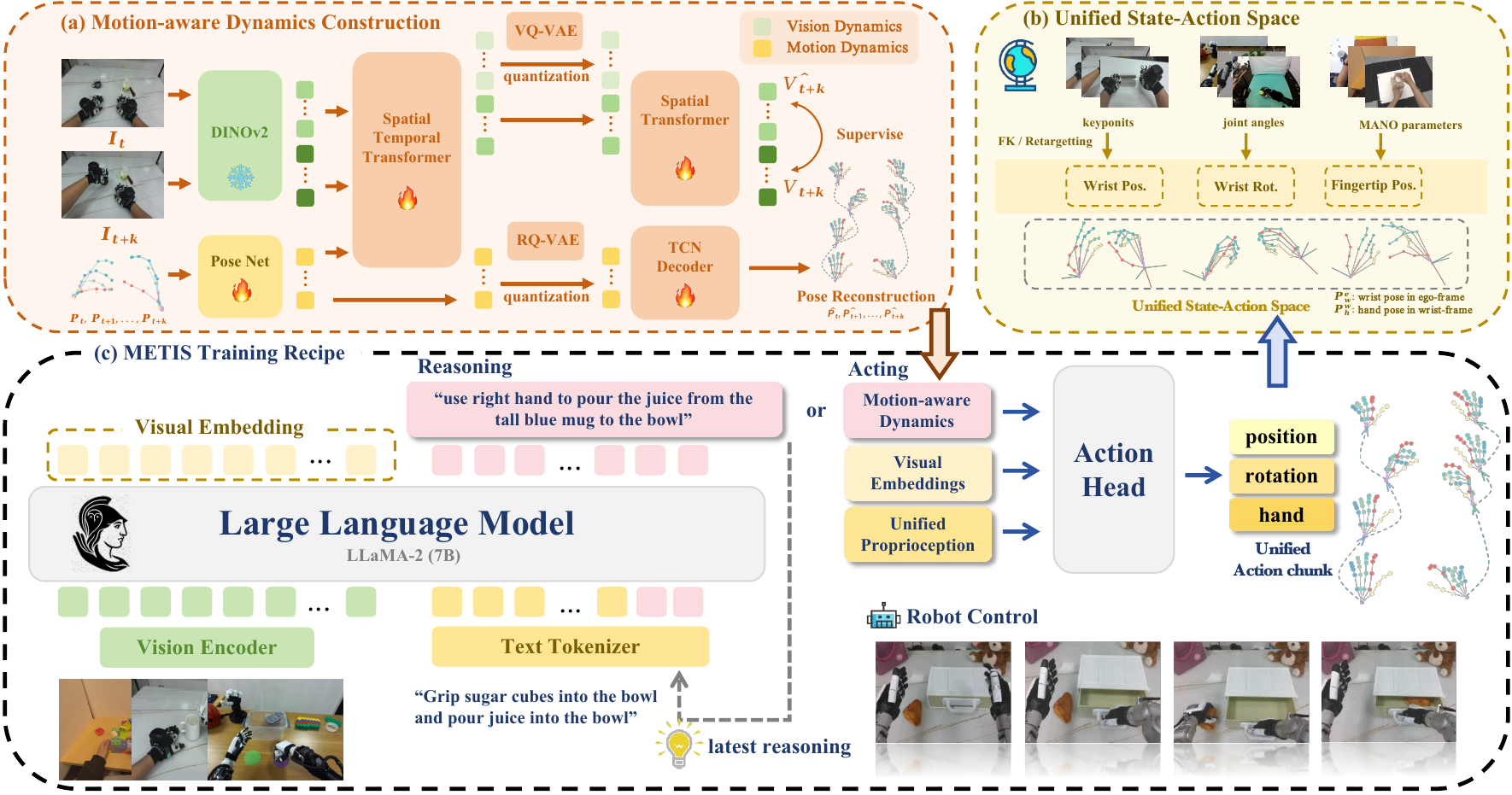}
    \caption{\textbf{Overview Framework} (a) We construct an expressive yet compact representation to capture the dynamics in dexterous manipulation. (b) METIS is pretrained on multi-source EgoAtlas dataset, where human and robot actions are align under a unified action space. (c) METIS integrates reasoning and acting within a framework, enabling effective deployment to downstream tasks.}
    \label{fig:pipeline}
    \vspace{-1em}
\end{figure*}
\subsection{Motion-Aware Dynamics Construction}
\label{Motion-Aware Dynamics Construction}
Recent VLA models typically employ generic tokenizers to discretize continuous actions, enabling autoregressive sequence modeling for policy learning. However, as the number of action chunks and system degrees of freedom increase, discrete action representations inevitably lead to longer token sequences, which in turn increases the computational cost and latency of autoregressive generation. Additionally, this tokenization approach often struggles to capture fine-grained motion details, which is particularly critical in dexterous manipulation, where subtle finger movements and contact interactions are essential.

To address these challenges, we propose a compact and discretized representation that effectively constructs motion-aware dynamics for dexterous manipulation. This representation provides an efficient and expressive supervision signal for VLA pretraining. The proposed dynamics model consists of two key components, detailed as follows.

\noindent\textbf{Visual Dynamics Discretization.}
A key challenge in egocentric dexterous manipulation is modeling how hand motion induces structured visual change. Instead of discretizing raw visual observations, we explicitly model \emph{motion-conditioned visual dynamics} that capture the causal effect of manipulation on future visual states.
Specifically, we employ an Inverse Dynamics Model based encoder $\mathcal{I}(D_{vis}|I_t, I_{t+k}, P_{t, t+1, ..., t+k})$ and a Forward Dynamics Model based decoder $\mathcal{F}(I_{t+k}|I_t, D_{vis})$. The encoder captures motion-relevant visual dynamics by integrating visual observations with continuous motion information, while the decoder is trained to predict future observation given the constructed visual dynamics. Our encoder includes both spatial and temporal transformer while the decoder only contains spatial transformer.

The visual dynamics $D_{vis}$ are quantized with the VQ-VAE objective \cite{van2017neural}. It maps continuous latent features into a finite set of discrete codebook embeddings, thereby enabling efficient and stable auto-regressive training for VLA. Following \citet{bu2025univla}, we do not reconstruct image pixels, as raw pixels  contain substantial redundancy and details irrelevant to the manipulation task. Instead, we use pretrained DINOv2 \cite{oquab2023dinov2} to extract high-level semantic representations, that better reflect task-relevant visual dynamics.

\noindent\textbf{Motion Dynamics Quantization.}
Learning motion priors is essential for VLA models. To this end, we introduce a separate set of discretized codebook embeddings $D_{mot}$ to focus on motion capturing. We discretize hand motion trajectories into compact dynamics tokens that effectively capture proprioceptive information and reflect the fine-grained features of dexterous manipulation. We propose PoseNet as the encoder for 3D hand motion, combining multi-scale temporal convolutions with trajectory self-attention to capture spatio-temporal dynamics. The continuous motion features are then quantized using RQ-VAE \cite{lee2022autoregressiveimagegenerationusing}, which not only effectively prevents codebook collapse but also captures hierarchical motion patterns, from coarse to fine levels. A temporal convolutional network (TCN) \cite{lea2017temporal} is used as the decoder to reconstruct motion trajectories during training, as illustrated in \Cref{fig:pipeline}(a). More details about motion-aware dynamics can be found in Appendix \ref{Appendix: Motion-aware  dynamics tokens}.

\subsection{METIS Model}
\label{VLA Architecture}
As shown in \Cref{fig:pipeline}(c), our proposed METIS model is built upon a VLM, where the parameters are initialized from the Prismatic-7B \cite{karamcheti2024prismatic}. It incorporates a hybrid vision encoder that integrates SigLIP \cite{zhai2023sigmoid} and DINOv2 \cite{oquab2023dinov2}, capturing both global semantics and fine-grained spatial details. The resulting visual features $f^{SigLIP}\in \mathbb{R}^{N_v \times 1152}$ and $f^{DINO}\in \mathbb{R}^{N_v \times 1024}$, where $N_v$ represents the visual token dimension, are concatenated along the channel dimension. A projection layer is applied to align visual embeddings with the language modality. The 7B LLaMA-2 \cite{touvron2023llama} large language model is adopted as the LLM backbone, which employs a decoder-only Transformer architecture with 32 sequential blocks for auto-regressive language modeling.

Previous work such as OpenVLA \cite{kim24openvla} and RT-2 \cite{zitkovich2023rt} map the infrequently used words in LLaMA-2 to action bins uniformly distributed within $[-1, 1]$. While effective for low-dimensional control, they struggle to scale to high-dimensional, continuous actions, often resulting in inefficient and unstable behavior. METIS first extends the LLaMA tokenizer vocabulary with $|C_1| + |C_2|$ special tokens, where $|C_1|$ and $|C_2|$ correspond to the codebook sizes of the visual dynamics tokens and motion dynamics tokens, respectively. Using the dynamics model introduced in the previous section, we discretize each egocentric manipulation sequence into motion-aware action tokens, denoted as $D_{vis}$ and $D_{mot}$. Each dynamics feature is assigned to its nearest codebook entry, and the resulting discrete index is mapped to a unique special token. This design fully leverages the original VLM architecture, enabling autoregressive supervision for training. By preserving the language priors of the language model while injecting motion information, METIS learns to model the fine-grained dynamics essential for dexterous manipulation. The auto-regressive objective of METIS $\pi$ is to minimize the sum of next-dynamics negative log-probabilities:
\vspace{-0.5em}
\begin{equation}
    \mathcal{L}_{ar} = \mathbb{E}_{o_t, l, a_{d,<i}}[-\sum_{i=1}^{N}log\; \pi_{\phi}(\hat{a}_{d,i|o_t,l,a_{d,<i}})]
\end{equation}
where $N$ represents the total length of dynamic tokens, which is 44 in our settings (4 for visual dynamics and 40 for motion dynamics). More details can be found in Appendix.

\noindent\textbf{Action Decoder.} The Action Decoder translates motion-aware dynamic tokens into executable low-level actions. It takes as input the dynamics token, visual embeddings, and the current proprioception expressed in the unified space. visual and dynamic features are aggregated through multi-head attention pooling, while proprioceptive inputs are projected into the hidden space via a two-layer MLP. The fused representation is passed through a linear projection head to predict a sequence of actions over one second (32 future steps). The final training objective is $\mathcal{L} = \mathcal{L}_{ar} + \lambda \mathcal{L}_{action}$.

\noindent\textbf{Chain-of-Thought Reasoning for Action.} Inspired by chain-of-thought prompting in large language models, METIS decomposes high-level manipulation instructions into shorter subtasks. The subtask is accompanied by fine-grained hand-level descriptions that explicitly guide action prediction (e.g., "grip sugar cubes into the bowl"). To support this reasoning process, we manually annotate our self-collected dataset with subtask-level labels. Each demonstration trajectory is segmented into a sequence of temporally contiguous subtask segments, each accompanied by a natural-language description specifying the manipulated object, the acting hand, and the intended motion. 

METIS operates in two distinct, adaptively switched modes: Reasoning Mode and Acting Mode. This dual-mode design allows the model to dynamically balance high-level subtask reasoning with low-level action execution, while avoiding unnecessary reasoning at every timestep and thereby reducing inference latency. Specifically, Mode switching is controlled by a token-gated mechanism. We introduce two special tokens: \emph{beginning of reasoning} ($[BOR]$) and \emph{beginning of dynamics} ($[BOD]$). When a subtask transition is detected, METIS first predicts $[BOR]$ and enters the reasoning mode, during which it auto-regressively performs subtask reasoning. The generated reasoning is incorporated into the language context for subsequent decisions.
Conversely, If the model deems reasoning unnecessary, such as continuing the execution of an ongoing subtask, it directly predicts the $[BOD]$ token to switch into the acting mode. Following this, METIS outputs motion-aware dynamics tokens which are decoded by the action decoder into low-level actions for execution.

To supervise the gate at each training step, we construct token targets from the annotated subtask boundaries. Inspired by the probabilistic interval supervision of OneTwoVLA~\cite{lin2025onetwovla}, we use a symmetric sampling schedule within a transition window of $[-10,+10]$ frames around each boundary $\tau_k$. The probability of assigning $[BOR]$ is
\begin{equation}
    p_{\mathrm{BOR}}(t)
    = 1 - \frac{|t-\tau_k|}{20},
    \qquad |t-\tau_k| \leq 10.
\end{equation}
This probability equals $1.0$ at the boundary and decreases linearly to $0.5$ at either end of the window. Frames within the window are assigned $[BOR]$ with probability $p_{\mathrm{BOR}}(t)$ and $[BOD]$ otherwise, while frames outside all transition windows are assigned $[BOD]$. Each transition therefore provides up to 21 candidate frames for $[BOR]$ supervision, rather than only a single boundary frame. 

\section{Experiments}

\subsection{Experiment Setup}
\noindent\textbf{Hardware Platform.} For hardware platform, we use a Unitree G1 humanoid robot equipped with a pair of Inspire 6-DoF dexterous hands for fine-grained manipulation. An Intel RealSense D435 camera is mounted on the robot’s head to capture egocentric RGB observations. 

\noindent\textbf{Self-collected Robot Data.} We collect robot demonstrations through human teleoperation with a glove-tracker system \cite{wang2024dexcap}. The tracker attached to the operator’s wrist captures precise wrist poses, which are converted into the robot arm’s joint configurations via inverse kinematics(IK). Simultaneously, a motion-capture glove records the fingertip trajectories, which are mapped to the dexterous hand joint space using an IK-based retargetting algorithm \cite{qin2023anyteleop}.

\noindent\textbf{Tasks.} We evaluated METIS on six dexterous manipulation tasks including three short-horizon and three long-horizon tasks, as shown in \Cref{fig:dexterous manipulation tasks}: (1) Pick and Place, (2) Close Laptop, (3) Open Drawer, (4) Grasp Two Drinks into Basket, (5) Put Cola into Basket, (6) Open Drawer and Put Bread. We further evaluate METIS on more fine-grained tool-use tasks (e,g. Pour Drinks and Water Flowers). Each task is collected with 100 high-quality demonstrations and is evaluated with 20 trials by default. 
We provide more details about these tasks in Appendix \ref{Appendix: task details}.

\begin{table*}[t]
    \renewcommand{\arraystretch}{1}
    \centering
    \caption{\textbf{Main results of six real-world tasks.} Each experiment is evaluated with 20 trials. SR denotes Success Rate, and PSR denotes Progress Success Rate.}
    \vspace{-0.5em}
    \label{tab:Main_Experiments}
    \resizebox{\linewidth}{!}{
    \begin{tabular}{c c c c | cc cc cc}
        \toprule
        \multirow{2}{*}{\bf Method} &
        \multicolumn{1}{c}{\bf Pick and Place} &
        \multicolumn{1}{c}{\bf Close Laptop} &
        \multicolumn{1}{c|}{\bf Open Drawer} &
        \multicolumn{2}{c}{\bf Grasp Two Drinks into Basket} &
        \multicolumn{2}{c}{\bf Put Cola into Basket} &
        \multicolumn{2}{c}{\bf Open Drawer and Put Bread} \\
        \cmidrule(lr){2-10}
        & \bf SR & \bf SR & \bf SR & \bf \hspace{18pt}SR & \bf PSR & \bf SR & \bf PSR & \bf \hspace{18pt}SR & \bf PSR \\
        \midrule
        ACT & 35.0\% & 65.0\% & \textbf{95.0\%} & \hspace{18pt}25.0\% & 40.0\% & 50.0\% & 53.3\% & \hspace{18pt}5.0\% & 5.0\% \\
        OpenVLA-OFT & 50.0\% & 80.0\% & 10.0\% & \hspace{18pt}40.0\% & 57.5\% & 55.0\% & 56.7\% & \hspace{18pt}0.0\% & 1.0\% \\
        $\pi_{0.5}$ & 60.0\% & 85.0\% & 70.0\% & \hspace{18pt}65.0\% & 72.5\% & \textbf{75.0\%} & \textbf{76.7\%} & \hspace{18pt}60.0\% & 65.0\% \\
        GR00T N1.5 & 70.0\% & 80.0\% & 80.0\% & \hspace{18pt}65.0\% & 70.0\% & 70.0\% & 73.3\% & \hspace{18pt}70.0\% & 72.5\% \\
        \rowcolor{gray!20} \bf METIS (Ours) & \textbf{85.0\%} & \textbf{95.0\%} & 90.0\% & \hspace{18pt}\textbf{75.0\%} & \textbf{85.0\%} & 70.0\% & \textbf{76.7\%} & \hspace{18pt}\textbf{75.0\%} & \textbf{82.5\%} \\
        \bottomrule
    \end{tabular}
    }
    \vspace{-0.7em}
\end{table*}

\begin{figure}[t]
    \centering

    \begin{minipage}[t]{0.52\linewidth}
        \centering
        \includegraphics[width=\linewidth]{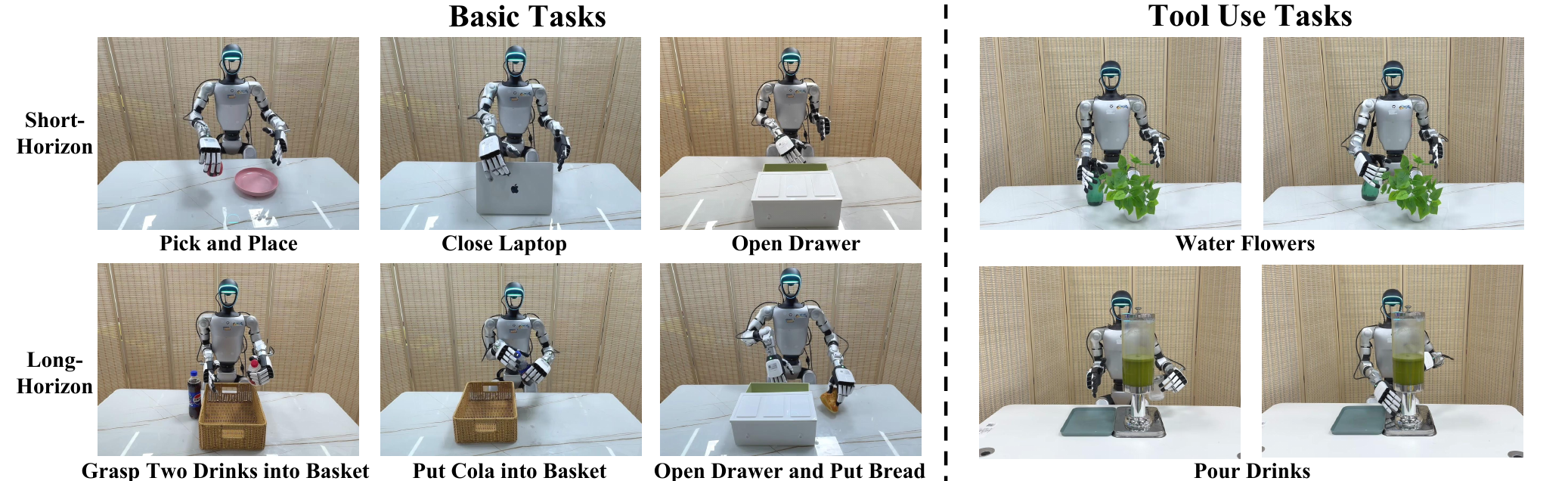}
        \vspace{-1.5em}
        \caption{\textbf{Visualization of dexterous tasks.}}
        \label{fig:dexterous manipulation tasks}
    \end{minipage}
    \hfill
    \begin{minipage}[t]{0.47\textwidth}
        \centering
        \vspace{-6em}
        \captionof{table}{\textbf{Performance on tool-use tasks.}}
        \label{tab:tool_use}
        \scriptsize
        \setlength{\tabcolsep}{7pt}
        \renewcommand{\arraystretch}{1.0}
        \begin{tabular}{c cc | cc}
            \toprule
            \multirow{2}{*}{\bf Method} &
            \multicolumn{2}{c|}{\bf Pour Drinks} &
            \multicolumn{2}{c}{\bf Water Flowers} \\
            & SR & PSR & SR & PSR \\
            \midrule
            $\pi_{0.5}$ & 40.0\% & 41.7\% & 15\% & 18.3\% \\
            GR00T N1.5 & 60.0\% & 75.0\% & 50\% & 53.3\% \\
            \rowcolor{gray!20} \bf METIS (Ours) & \textbf{80.0\%} & \textbf{83.3\%} & \textbf{60\%} & \textbf{63.3\%} \\
            \bottomrule
        \end{tabular}
    \end{minipage}

    \vspace{-1em}
\end{figure}

\noindent\textbf{Baselines.} We compare METIS with four representative baselines: (1) ACT \cite{zhao2023learning}, an action chunking transformer that learns low-level visuomotor policy. (2) OpenVLA-OFT \cite{kim2025fine}, an enhanced open-source VLA model via optimized fine-tuning; (3) $\pi_{0.5}$ \cite{intelligence2025pi_}, a VLA flow model for general robot control; (4) GR00T N1.5 \cite{bjorck2025gr00t}, an improved VLA model for generalist humanoid robots.

\noindent\textbf{Evaluation Metrics.} We use two metrics to evaluate model performance: Success Rate (SR), indicating the entire task is successfully completed, and Progress Rate (PSR), capturing the average completion ratio of subtasks relative to the overall task in long-horizon settings.

\subsection{Performance in Real-World Experiments} 
\label{Performance in Real-World Experiments}

We conduct experiments on real-world dexterous manipulation tasks, as shown in \Cref{tab:Main_Experiments}. Our model, achieves the highest average success rate and consistently outperforms existing state-of-the-art VLA models on most tasks. The specialist model ACT performs well on short-horizon tasks but struggles on long-horizon tasks. $\pi_{0.5}$ underperforms on tasks requiring precise dexterous manipulation, as it has not been pretrained on large-scale dexterous datasets. GR00T N1.5 achieves competitive results through large-scale pretraining on humanoid robot and human data with implicit latent representations. Nevertheless, the absence of explicit reasoning and feedback correction limits its performance on long-horizon tasks. METIS,  benefiting from motion-aware dynamics and explicit reasoning mechanism, achieves outstanding performance across both short and long-horizon tasks. Additionally, METIS attains highest PSR across all long-horizon tasks, demonstrating it can reason and act coherently across long-horizon sequences, maintaining stable control and minimizing error accumulation during task execution.

\noindent\textbf{Dexterous Capability.} We evaluate model performance on tool-use tasks to better assess fine-grained dexterous manipulation capability. As shown in \Cref{tab:tool_use}, dexterous VLA (e.g., GR00T N1.5 and METIS) consistently outperform the gripper-based VLA method ($\pi_{0.5}$). 
Among them, METIS achieves the best overall performance, benefiting from large-scale pretraining on human tool-use demonstrations and the extraction of compact dexterous motion representations.


\begin{table}[t]
    \centering
    \begin{minipage}[t]{0.48\textwidth}
        \centering
        \caption{\textbf{Ablation on multi-source pretraining.}}
        \label{tab:Ablation on multi-source egocentric pretraining}
        \scriptsize
        \setlength{\tabcolsep}{3pt}
        \begin{tabular}{c c c}
            \toprule
            \bf Method & \bf Pick and Place & \bf Put Bread \\
            \midrule
            METIS-NoPretrain & 60.0\% & 35.0\% \\
            METIS-HumanPretrain & 70.0\% & 60.0\% \\
            METIS-HumanRobotPretrain & 80.0\% & 65.0\% \\
            \rowcolor{gray!20} METIS-FullPretrain & \textbf{85.0\%} & \textbf{75.0\%} \\
            \bottomrule
        \end{tabular}
    \end{minipage}
    \hfill
    \begin{minipage}[t]{0.48\linewidth}
        \centering
        \captionof{table}{\textbf{Ablation on reasoning mechanism.}}
        \label{tab:Ablation on Reasoning Mechanism}
        \scriptsize
        \setlength{\tabcolsep}{3pt}
        \begin{tabular}{cccc}
        \toprule
        \multirow{2}{*}{\bf Method} &
        \bf Open Drawer and &
        \bf Pour & \bf Throughput\\
        & \textbf{Put Bread} & \textbf{Drinks} & \textbf{(Hz)}\\
        \midrule
        No-reasoning & 70.0\% & 65.0\% & 28.30 \\
        Always-reasoning & \textbf{75.0\%} & 75.0\% & 17.27 \\
        \rowcolor{gray!20}Adaptive-reasoning & \textbf{75.0}\% & \textbf{80.0\%} & \textbf{23.58} \\
        \bottomrule
        \end{tabular}
    \end{minipage}
    \vspace{-1em}

\end{table}

\begin{figure}[t]
    \centering

    \begin{minipage}[t]{0.49\linewidth}
        \centering
        \includegraphics[width=\linewidth]{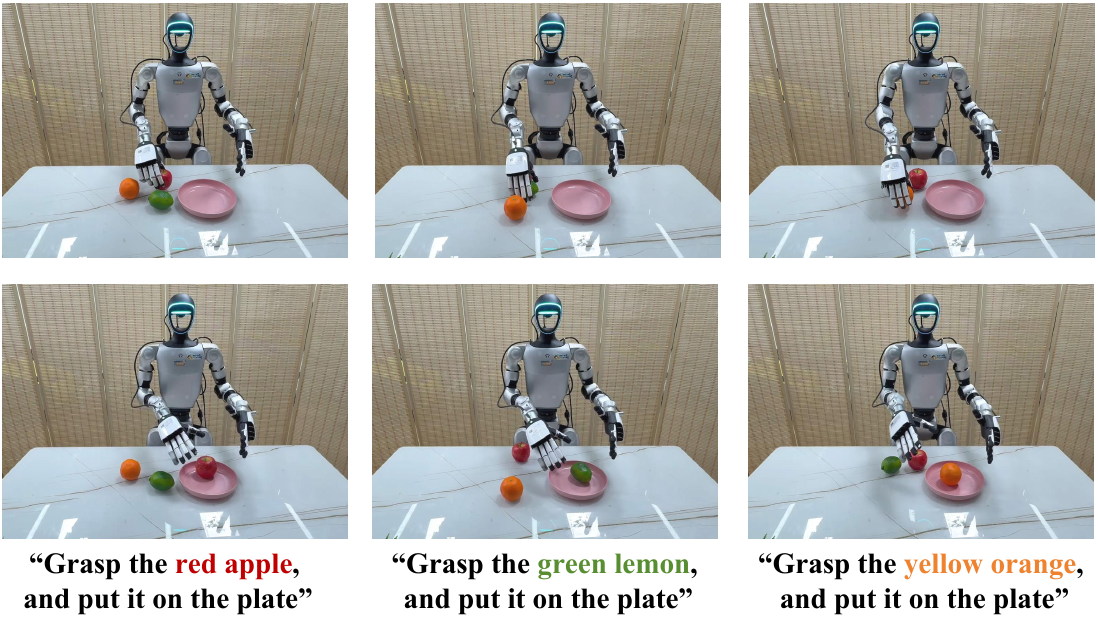}
        \caption{\textbf{Instruction following results.} Each task is collected with 100 demonstrations, jointly trained, and evaluated using different instructions.}
        \label{fig:instruction following}
    \end{minipage}
    \hfill
    \begin{minipage}[t]{0.49\linewidth}
        \centering
        \includegraphics[width=\linewidth]{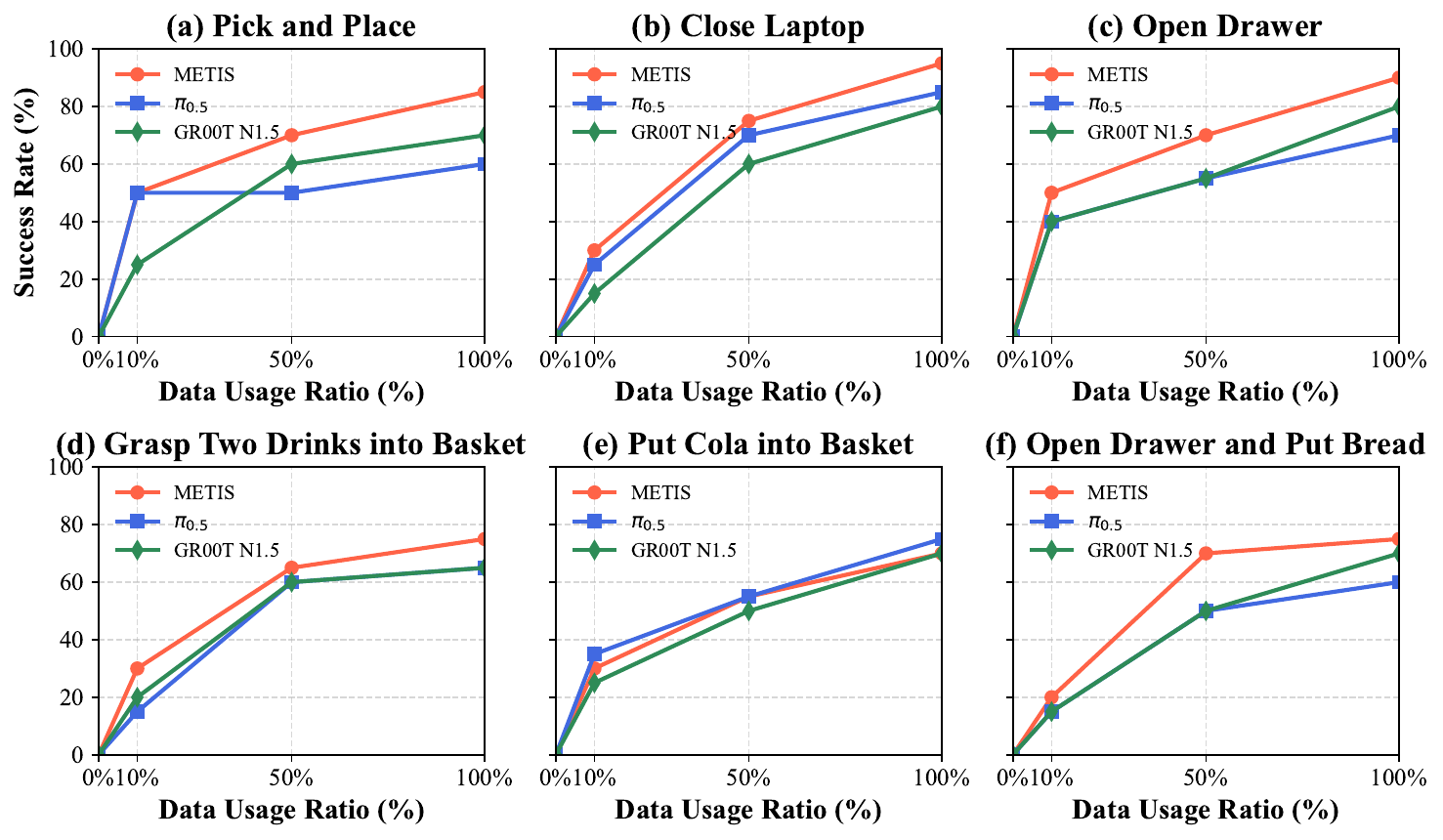}
        \caption{\textbf{Experimental results of efficiency. }We train the VLA model with an increasing number of demonstrations.}
        \label{fig:efficiency}
    \end{minipage}
    \vspace{-1em}

\end{figure}

\noindent\textbf{Instruction Following.} We further evaluate the instruction following capability of METIS. In this setup, three fruits of different colors (apples, orange, and lemon) are placed on the table. As shown in \Cref{fig:instruction following}, when given different language instructions (e.g., "place the red apple on the plate"), METIS can identify the target fruit and execute the corresponding grasping action. These results demonstrate the ability of METIS to connect object-level language understanding with dexterous manipulation in scenes containing multiple candidate objects.

\subsection{Efficiency}
\label{Efficiency}

\noindent\textbf{Sample Efficiency.} We evaluate the sample efficiency of METIS by training it with varying amounts of data on downstream dexterous tasks, as shown in \Cref{fig:efficiency}. METIS exhibits excellent efficiency, achieving superior performance even with limited training data. Notably, when fine-tuned with only 10\% data, METIS attains a 50\% success rate on the Pick and Place and Open Drawer task. This results strongly indicate that pretraining on diverse multi-source egocentric data with unified action space provides valuable priors, including spatial reasoning, visual-hand coordination, and motion dynamics. Such priors form consistent visuomotor representations and facilitate rapid transfer to new downstream tasks.

\noindent\textbf{Inference Efficiency.} Motion-aware dynamics provide a compact representation of hand motion, significantly reducing the number of tokens generated by the VLM and thereby improving inference efficiency. We evaluate METIS against other tokenizers (Naive and Fast \cite{pertsch2025fast}) under identical settings on a single NVIDIA H100 GPU. As shown in \Cref{tab:Inference Efficiency}, when predicting an action chunk of length 32, METIS with motion-aware dynamics reduces the token count from 1440 and 138 to only 44 tokens, achieving a throughput of 28.30~Hz (action steps per second). This demonstrates the effectiveness of motion-aware dynamics for efficient autoregressive generation.

\begin{table}[t]
    \centering
    \caption{\textbf{Ablation on motion-aware dynamics (success rate vs. inference efficiency).}}
    \label{tab:Inference Efficiency}
    \scriptsize
    
    \resizebox{\linewidth}{!}{
    \begin{tabular}{c c c c c c}
        \toprule
        \bf Method &
        \multicolumn{1}{c}{\bf Token Count} &
        \multicolumn{1}{c}{\bf Throughput (Hz)} &
        \bf Pick and Place & 
        \bf Open Drawer and Put Bread & 
        \bf Pour Drinks
        \\
        \midrule
        METIS w/ Naive Tokenizer &  1440 & 0.93 & 15.0\% & 10.0\% & 0.0\% \\
        METIS w/ Fast Tokenizer &  138 & 9.87 & 50.0\% & 25.0\% & 15.0\%  \\
        \rowcolor{gray!20} METIS w/ Motion-aware Dynamics & \textbf{44} & \textbf{28.30} & \textbf{60.0\%} & \textbf{35.0\%} & \textbf{30.0\%} \\
        \bottomrule
    \end{tabular}
    }
    \vspace{-2em}
\end{table}


\subsection{Generalization}
\label{Generalization}
Besides the remarkable effectiveness and efficiency, METIS also showcases excellent generalization capabilities. We evaluate our model in four \textit{out-of-distribution}(OOD) scenarios, including (1) \textbf{Unseen background}, we cover the tabletop with a colorful patterned tablecloth to introduce significant visual distractions and background textures unseen during training. (2) \textbf{Unseen lighting condition}, we illuminate the scene with color-changing and flickering lights to simulate dynamic and diverse lighting situations. (3) \textbf{Unseen object}, we replace the original scone bread with a visually distinct croissant to assess object-level generalization. (4) \textbf{Cluttered environment}, we randomly place distractor objects such as a plate and an apple near the drawer, increasing visual and spatial complexity. Taking the Open Drawer and Put Bread task as an example, we compare and report the success rates of METIS and GR00T N1.5 across all four out-of-distribution scenarios, as shown in \Cref{tab:Generalization}. The results demonstrate that METIS effectively adapts to various distributional shifts, maintaining stable visuomotor grounding and task execution even under significant visual or physical variations.


\begin{table}[t]
    \centering

    \begin{minipage}[t]{0.52\textwidth}
        \centering

        \captionof{table}{\textbf{Generalization results of OOD scenarios.} }
        \includegraphics[width=0.85\linewidth]{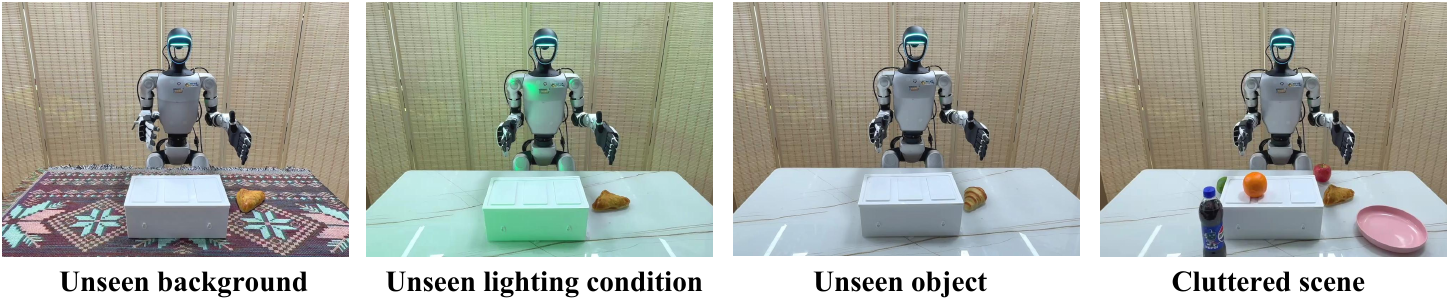}
        \label{tab:Generalization}
        
        \scriptsize
        \setlength{\tabcolsep}{3pt}
        \begin{tabular}{c c c c c}
            \hline
            \bf Method & \bf Background & \bf Lighting & \bf Object & \bf Cluttered \\
            \hline
            $\pi_{0.5}$ & 50.0\% & \textbf{70.0\%} & 65.0\% & 55.0\% \\
            GR00T N1.5 & 65.0\% & 65.0\% & 65.0\% & 60.0\% \\
            \rowcolor{gray!20} \bf METIS (Ours) & \textbf{70.0\%} & 65.0\% & \textbf{70.0\%} & 65.0\% \\
            \hline
        \end{tabular}
    \end{minipage}
    \hfill
    \begin{minipage}[t]{0.46\textwidth}
        \centering
        \vspace{1em}
        \includegraphics[width=\linewidth]{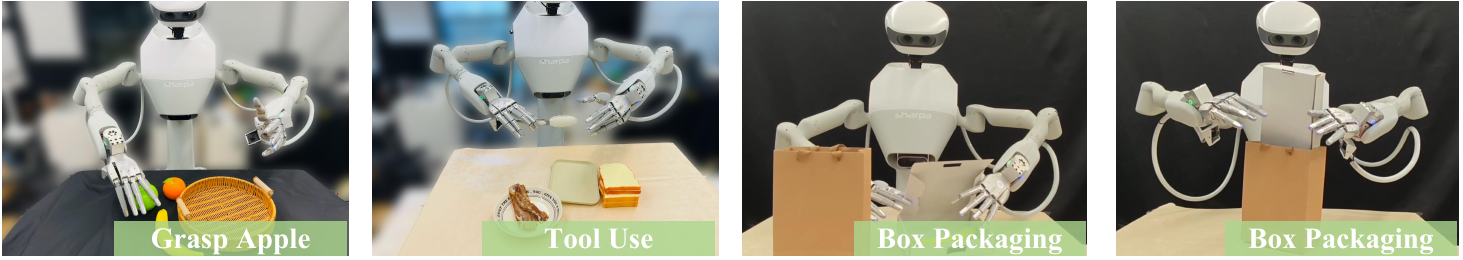}
        \vspace{-0.8em}
        \captionof{figure}{\textbf{Cross-embodiment generalization.} METIS demonstrates transferability to 22-DoF dexterous hands, achieving stable performance on three dexterous tasks.}
        \label{fig:cross-embodiment generalization}
    \end{minipage}
    \vspace{-1em}
\end{table}

\noindent\textbf{Cross-Embodiment Generalization.} METIS also generalizes effectively to higher-DoF dexterous hands, demonstrating strong adaptability across embodiments. We evaluate METIS on the Sharpa Beta embodiment equipped with a pair of 22-DoF SharpaWave Dexterous hands, as shown in \Cref{fig:cross-embodiment generalization}. The model achieves 85.0\% success rate on the Grasp Apple into Basket task and 70.0\% on the Tool Use task, respectively. We further evaluate METIS on a long-horizon and highly dexterous \emph{Box Packaging} task, which requires precise grasping, coordinated bimanual manipulation, folding of packaging materials, and accurate placement into a box. Despite the increased task complexity, METIS achieves a 15\% success rate, while the baseline GR00T N1.5 fails to complete the task in all 20 trials. As METIS predicts fingertip trajectories rather than direct joint angles, 
the policy is naturally transferable and remains unaffected by variations in hand kinematics.


\subsection{Ablations}
\label{Ablations}
We have demonstrated the effectiveness of METIS in terms of overall performance, efficiency, and generalization. However, the contribution of its core components remains unexplored. In this section, we perform ablation studies focusing on three aspects: (1) the effect of multi-source egocentric pretraining, (2) the impact of the motion-aware dynamics, (3) the effect of reasoning mechanism, and (4) the visual dynamics token budget.

\noindent\textbf{Multi-source Egocentric Pretraining.} 
To investigate the role of multi-source egocentric pretraining, we evaluate four variants of the model on downstream tasks: (a) METIS without pretraining, (b) METIS pretrained only on the open-source human data, (c) METIS pretrained on the human and robot data, excluding our self-collected dataset, (d) METIS pretrained on EgoAtlas dataset. As presented in \Cref{tab:Ablation on multi-source egocentric pretraining}, pretraining on multi-source egocentric datasets improves downstream performance, indicating that training on diverse visual and action distributions enables the model to learn more robust visuomotor priors and generalize effectively across a wide range of tasks. Notably, although the no-pretraining VLA achieves a certain level of success rate on some tasks, it exhibits significant loss fluctuations during the post-training stage and unstable joint jitters during real-world deployment.


\noindent\textbf{Motion-aware Dynamics.} We conduct ablation experiments on the motion-aware dynamics. During post-training, we compare motion-aware dynamics with naive and FAST tokenizers under the same setting without pretraining. As shown in \Cref{tab:Inference Efficiency}, METIS with motion-aware dynamics consistently achieves better performance, especially on long-horizon and fine-grained tasks, highlighting its advantage in modeling motion details. 

To assess the contribution of each component, we compare visual-only, motion-only, and full motion-aware dynamics on two manipulation tasks. As shown in \Cref{tab:visual_dynamics_ablation}(a), the full formulation achieves success rates of 75.0\% on Open Drawer and Put Bread and 80.0\% on Pour Drinks, outperforming both visual-only dynamics (45.0\% and 50.0\%) and motion-only dynamics (60.0\% and 55.0\%). These results support the complementary roles of the two components: visual dynamics models scene evolution and object-state transitions, while motion dynamics directly supervises wrist and fingertip kinematics. We provide more quantitative results about codebook usage, motion reconstruction and visualizations of motion-aware tokens, please refer to Appendix \ref{Quantitative Results and Visualization}.

\noindent\textbf{Reasoning Mechanism.} We conduct a quantitative evaluation of the reasoning mechanism by comparing three variants: no-reasoning, always-reasoning, and our adaptive-reasoning. The success rate and inference latency are reported in \Cref{tab:Ablation on Reasoning Mechanism}. For long-horizon dexterous tasks, reasoning improves planning capability and overall performance. While always-reasoning achieves competitive success rates, it incurs higher latency due to generating both reasoning and actions at every step, which can be detrimental in highly dynamic settings.

\noindent\textbf{Visual Dynamics Token Budget.}  
We further investigate the visual-token budget through codebook usage and visual reconstruction error. As reported in \Cref{tab:visual_dynamics_ablation}(b), increasing the budget from four to eight tokens reduces reconstruction error from 6.0 to 2.4, while codebook usage decreases slightly from 1.0 to 0.95. Although additional tokens improve reconstruction fidelity within this range, they also increase the sequence length and associated decoding cost. We adopt four visual dynamics tokens as a practical trade-off between reconstruction fidelity and inference efficiency. Qualitative visualizations in Appendix \Cref{Visualization of Codewords} further show that similar motion patterns, such as downward or rightward movements, can be represented by the same codewords across different embodiments and scene appearances. This suggests that the learned visual dynamics tokens capture motion semantics that are partially invariant to embodiment and visual context.


\begin{table}[t]
    \centering
    \caption{\textbf{Ablations of visual dynamics.}
    (a) Success rates with different dynamics components.
    (b) Effect of the visual-token budget on codebook usage
    and visual reconstruction error.}
    \label{tab:visual_dynamics_ablation}

    \begin{minipage}[t]{0.53\linewidth}
        \vspace{0.5pt}
        \centering
        \scriptsize
        \textbf{(a) Dynamics components}\\[0.5em]
        \setlength{\tabcolsep}{3pt}
        \renewcommand{\arraystretch}{1.3}

        \begin{tabular}{@{}lcc@{}}
            \toprule
            \textbf{Method}
            & \shortstack{\textbf{Open Drawer}
                          \textbf{and Put Bread}}
            & \shortstack{\textbf{Pour}
                          \textbf{Drinks}} \\
            \midrule
            Visual-only
            & 45.0\% & 50.0\% \\
            Motion-only
            & 60.0\% & 55.0\% \\
            \rowcolor{gray!20}
            \textbf{Motion-aware}
            & \textbf{75.0\%} & \textbf{80.0\%} \\
            \bottomrule
        \end{tabular}
    \end{minipage}
    \hfill
    \begin{minipage}[t]{0.45\linewidth}
        \vspace{0.5pt}
        \centering
        \scriptsize
        \textbf{(b) Visual-token budget}\\[0.5em]
        \setlength{\tabcolsep}{4pt}
        \renewcommand{\arraystretch}{0.95}

        \begin{tabular}{@{}ccc@{}}
            \toprule
            \shortstack{\textbf{Visual}
                        \textbf{Tokens}}
            & \shortstack{\textbf{Codebook}
                          \textbf{Usage}}
            & \shortstack{\textbf{Reconstruction}
                          \textbf{Error} $\downarrow$} \\
            \midrule
            2 & 0.80 & $1.2e^{-2}$ \\
            4 & 1.00 & $6.0e^{-4}$ \\
            6 & 0.95 & $3.9e^{-4}$ \\
            8 & 0.95 & $2.4e^{-4}$ \\
            \bottomrule
        \end{tabular}
    \end{minipage}
    \vspace{-2em}
\end{table}


\vspace{-0.5em}
\section{Conclusion} 
\label{sec:conclusion}
In this paper, we present METIS, a vision-language-action model for dexterous manipulation pretrained on multi-source egocentric dataset, integrating reasoning and acting within a unified framework. To support large-scale pretraining, we construct EgoAtlas, a comprehensive multi-source egocentric dataset that aligns human and robotic data under a consistent action space. By extracting motion-aware dynamics from manipulation trajectories, METIS acquires compact and expressive representations of hand motion, enabling precise and coordinated action generation. Experimental results demonstrate that METIS achieves strong performance across diverse dexterous manipulation tasks and exhibits robust generalization to out-of-distribution scenarios.


\begin{ack}
This work was supported by the National Natural Science Foundation of China (62476011) and Beijing Natural Science Foundation (L252060). We thank BAAI for providing hardware and computational resources for our experiments. We are also grateful to Lei Zhang and Shaoxuan Xie for their assistance with data collection.
\end{ack}

\bibliographystyle{plainnat}
\bibliography{references} 


\clearpage
\appendix

\noindent{\large\bfseries Appendix}
\section{EgoAtlas Dataset}
\subsection{\textbf{Data Calibration}}
We perform a comprehensive calibration procedure using three VIVE trackers within our motion-capture setup. Two trackers are mounted on the wrists to provide the 6-DoF global wrist poses in the world coordinate frame. Another tracker is rigidly attached to the head-mounted egocentric camera via a custom-designed mount. Given this fixed transformation, we obtain the extrinsic parameters of the egocentric camera with respect to the global coordinate frame. By combining these measurements, we transform the wrist poses from the world coordinate frame into the egocentric camera coordinate frame, achieving consistent spatial correspondence between hand motion and egocentric observations.
For the finger poses, the Manus Quantum gloves directly provide 3D fingertip positions relative to the wrist frame.

\subsection{\textbf{Data Collection Interface}} 
To ensure data quality, we built a data collection interface that visualizes the video stream, wrist pose and fingertip keypoints in real time. This platform allows the operator to monitor the tracking status of all sensors. When the operator identifies tracking failures or degraded quality, the recording session can be manually terminated anytime. The entire system operates at 20Hz, balancing motion fidelity with reliable multi-sensor synchronization.

\subsection{\textbf{Coordinate System Definition}} 
\label{Appendix: Coordinate System Definition}
The egocentric camera coordinate system follows the conventional computer vision convention: the origin is located at the optical center of the camera, the z-axis points forward along the optical axis (outward from the camera), the x-axis points to the right of the camera, and the y-axis points downward in the image plane.

When computing the fingertip positions of the dexterous hand, we employ forward kinematics (FK) with the root joint of the hand as the origin. However, the base of the robotic hand is typically not perfectly aligned with the wrist pose. To resolve this, we apply a fixed transformation between the wrist and the hand base to ensure geometric consistency. For clarity, we define wrist coordinate systems as follows:
\begin{itemize}
    \item \textbf{Left Hand}: the \textbf{x-axis} points toward the fingertips, the \textbf{y-axis} from the palm to the back of the hand, and the \textbf{z-axis} from the little finger toward the thumb.
    \item \textbf{Right Hand}: the \textbf{x-axis} points along the forearm direction, the \textbf{y-axis} from the palm to the back of the hand, and the \textbf{z-axis} from the little finger toward the thumb.
\end{itemize}

\begin{figure}[h]
    \centering
    \includegraphics[width=\linewidth]{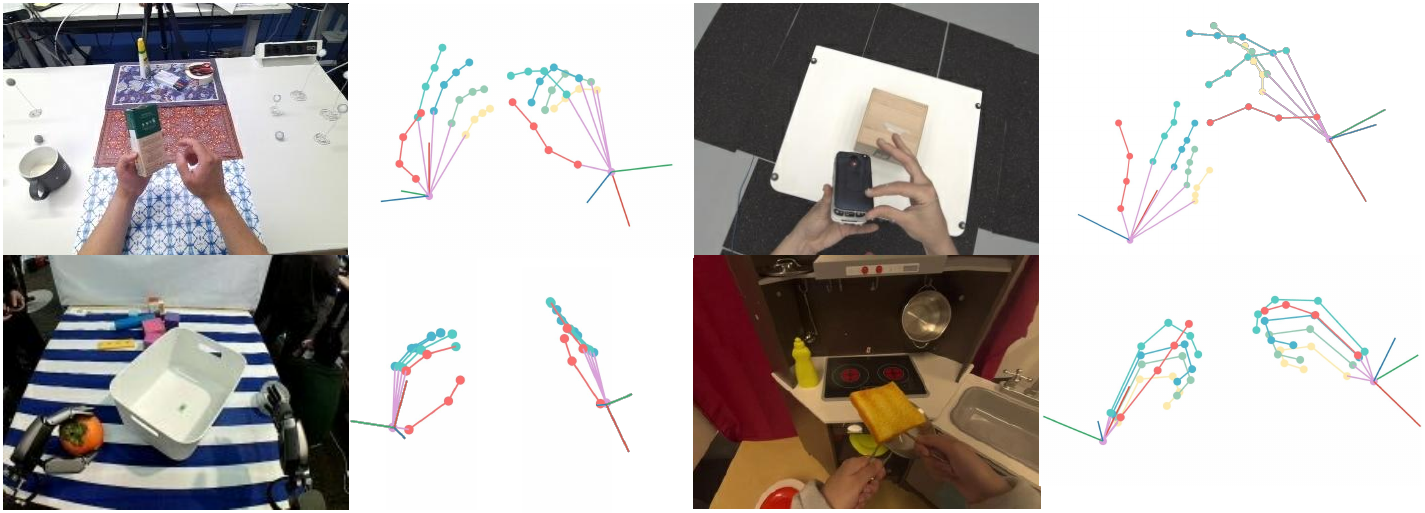}
    \caption{\textbf{Visualization of hand motion.} We use 3D keypoints to represent hand motion.  To ensure consistency across embodiments, the wrist frames of human and robot hands are strictly aligned.}
    \label{fig:coordinate}
    \vspace{-1em}
\end{figure}

We provide a visualization of the hand motion in our dataset (see \Cref{fig:coordinate}).

\subsection{\textbf{Unifying Action Space}} 
\label{Appendix: Unifying Action Space}
In this section, we describe how to map hand motion information from different configurations and embodiments into a unified action space. Existing action representations for hand motion can generally be categorized into three classes.

\begin{itemize}
    \item \textbf{MANO-based human hand representations.} 
    \item \textbf{Keypoint-based hand representations.}
    \item \textbf{Robot joint-space representations.}
\end{itemize}

\textbf{MANO-based representations model human hand motion} using a low-dimensional parametric hand model, where pose and shape are encoded by a set of continuous parameters. Such representations are typically collected via vision-based motion capture systems \cite{fan2023arctic, kwon2021h2o, wang2023holoassist, yang2022oakink, liu2022hoi4d}, including multi-view RGB cameras or RGB-D sensors, which estimate hand pose by fitting the MANO model to visual observations. During data processing, the MANO pose parameters are first passed through the parametric hand model to recover 3D joint locations (MANO FK). A subset of anatomically meaningful position, corresponding to the fingertips of each finger, is then selected to represent fine-grained hand motion. To decouple global hand motion from local finger configurations, all fingertip positions are expressed in a wrist-centric coordinate frame by removing the root rotation and translation.

In parallel, the wrist pose is extracted from the MANO root parameters and represented as a 6D rigid transformation. Using known camera extrinsics, the wrist pose is further transformed from the world coordinate frame into the egocentric camera frame. The resulting wrist pose and fingertip positions jointly form an explicit, viewpoint-consistent spatial representation, which serves as a unified state-action representation for policy learning.

\textbf{{Keypoint-based hand representations}} using explicit 3D joint positions, typically acquired through VR systems \cite{hoque2025egodex} or motion-capture gloves. In these settings, egocentric visual observations are typically provided by a head-mounted VR device or an external camera, while head and wrist poses are obtained through VR systems or dedicated motion trackers. During data processing, hand and finger joint poses are initially provided in a global or world coordinate frame. Using the calibrated camera extrinsics, all joint poses are first transformed into the egocentric camera frame, ensuring alignment with the visual observations. A subset of finger keypoints corresponding to fingertip endpoints is selected to capture fine-grained hand articulation, and these points are expressed in wrist-local coordinates by removing the wrist translation and rotation.

\textbf{Robot joint-space representations.} Robotic arms and hands usually represent actions in joint space \cite{fourier2025actionnet}, where each action corresponds to target joint angles. Given the robot kinematic model defined by its URDF, forward kinematics can be applied to recover the end-effector pose relative to the robot base frame. Similarly, forward kinematics over the hand kinematic chain enables the computation of fingertip positions expressed in the wrist coordinate frame. To ensure consistency with egocentric perception, the end-effector pose is further transformed into an egocentric coordinate frame using the calibrated extrinsic parameters of the head-mounted camera.

\subsection{\textbf{Subtask-Level Annotation}} 
\label{Appendix: Subtask-Level Annotation}
To enable fine-grained understanding of long-horizon manipulation behaviors, we annotate each demonstration video with subtask-level temporal labels. Each video is decomposed into a sequence of semantically meaningful atomic actions, where each subtask corresponds to a continuous temporal segment associated with a natural language description.

Annotators manually partition the video into a set of non-overlapping temporal intervals. Each interval is defined by a start frame and an end frame, forming a contiguous segment that captures a coherent manipulation primitive. The segmentation strictly preserves the temporal order of the task execution and covers the entire video without gaps.

\begin{figure*}[h]
    \centering
    \includegraphics[width=\linewidth]{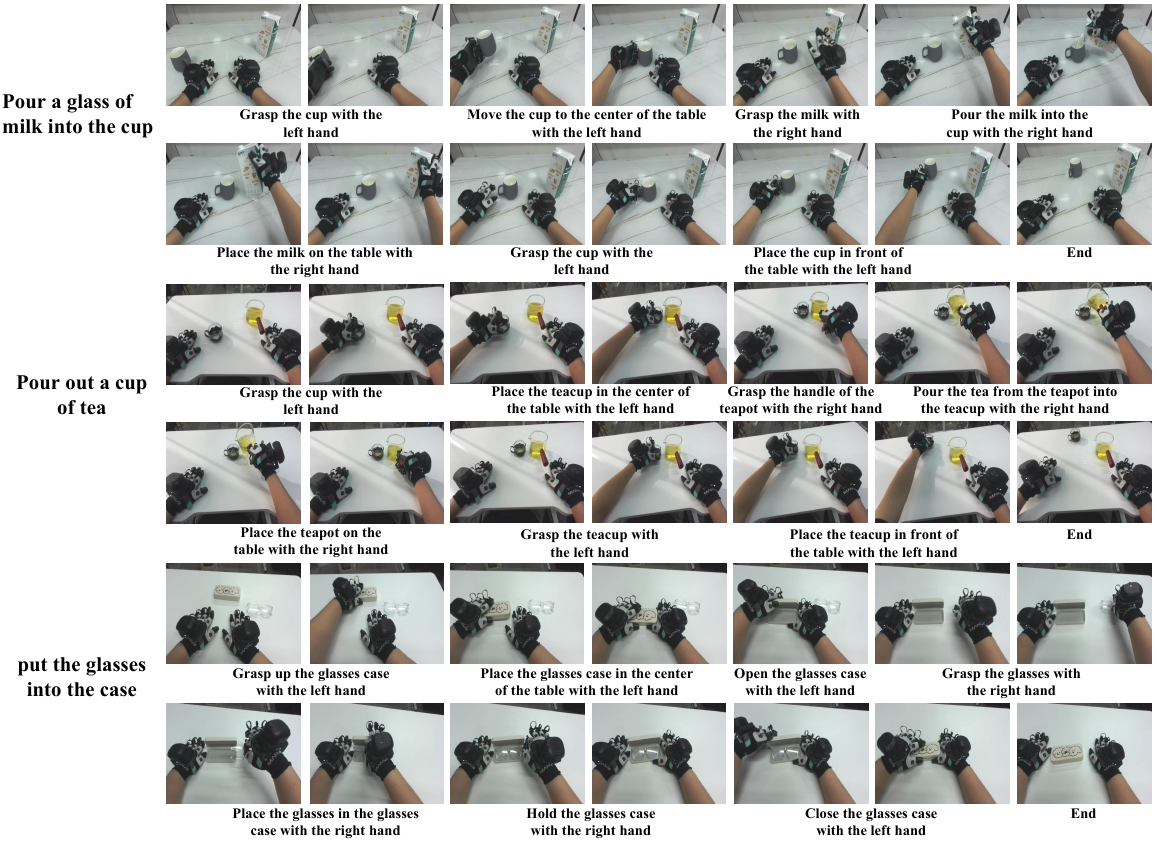}
    \caption{\textbf{Visualization of subtask-level annotations.}}
    \label{fig:subtast annotation}
\end{figure*}

For each temporal segment, annotators assign a concise description that specifies:
\begin{itemize}
    \item The action primitive (e.g., grasp, move, pour, place).
    \item The active effector (left hand or right hand).
    \item The target object when applicable.
\end{itemize}
Each episode concludes with an explicit terminal label (e.g., End) to mark task completion. This provides a clear boundary for long-horizon reasoning and facilitates sequence-level modeling. Figure \ref{fig:subtast annotation} visualizes the subtask-level annotations of representative data samples.

\subsection{Hand IK Retargetting}
\label{Appendix: Hand IK Retargetting}
We use Pinocchio to parse the hand URDF and perform kinematic computations, including forward kinematics and fingertip Jacobian evaluation, which are then used in our iterative IK solver. Given the hand URDF and predefined master–mimic relations, the retargeting pipeline is deterministic and geometry-based, and consistently applicable across embodiments (validated by cross-embodiment results). 
The IK solver may introduce approximation errors, especially for high-DoF dexterous hands where redundancy, joint limits, and embodiment-specific kinematic constraints can lead to multiple feasible solutions or no exact solution.  To mitigate this issue, our solver is initialized from the current hand state and optimized with joint-limit and smoothness constraints, which biases the solution toward physically feasible and temporally consistent motions. When an exact fingertip target is unreachable, the solver returns the closest stable solution rather than an arbitrary pose, avoiding abrupt joint changes and reducing self-collisions.
Any residual projection error is further absorbed during embodiment-specific fine-tuning, where the policy adapts to the reachable workspace and kinematic constraints of the target hand. 

\subsection{Data Sources and Statistics}
\label{Appendix: Data Sources and Statistics}
\begin{table}[b]
    \centering
    \caption{\textbf{Statistics of EgoAtlas.} We use in-the-wild to denote data collected in diverse, unconstrained real-world scenes.}
    \label{tab:Statistics of EgoAtlas}
    \resizebox{\linewidth}{!}{
        \begin{tabular}{c|cccccccc}
            \toprule
            \bf{Method} & \bf{Trajs} & \bf{Frames} & \bf{Hours} & \bf{Pose} & \bf{Subtask} & \bf{Human} & \bf{Robot} & \bf{In-the-wild}\\
            \midrule
            {ARCTIC}
            &  {296} & {214.5K} & {2.0} & \ding{51} & \ding{55} & {100\%} & {0\%} & \ding{55}\\
            {H2O}
            &  {109} & {65.3K} & {0.6} & \ding{51} & \ding{51} & {100\%} & {0\%} & \ding{55} \\
            {HoloAssist}
            & {100} & {777.3K} & {7.3} & \ding{51} & \ding{51} & {100\%} & {0\%} & \ding{55} \\
            {Oakink}
            & {134} & {146K} & {1.4} & \ding{51} & \ding{51} & {100\%} & {0\%} & \ding{55} \\
            {EgoDex}
            & {314.8K} & {77.9M} & {721.3} & \ding{51} & \ding{55} & {100\%} & {0\%} & \ding{51}\\
            {PH2D}
            & {1.8K} & {416.5K} & {3.9} & \ding{51} & \ding{55} & {66.1\%} & {33.9\%} & \ding{51} \\
            {ActionNet}
            & {15.7K} & {7.4M} & {68.4} & \ding{51} & \ding{55} & {0\%} & {100\%} & \ding{51} \\
            {Ours}
            &  {10K} & {2.8M} & {38.9} & \ding{51} & \ding{51} & {100\%} & {0\%} & \ding{51} \\
            \bottomrule
        \end{tabular}
    }
\end{table}

EgoAtlas integrates data from four major sources, covering both human and robotic domains with diverse sensing modalities: \textbf{(1) Vision-based motion capture datasets} \cite{fan2023arctic, kwon2021h2o, wang2023holoassist, yang2022oakink, liu2022hoi4d}, which use multi-camera optical systems to capture precise 3D hand annotations and object interactions. However, they are typically confined to small tabletop environments with limited diversity.
\textbf{(2) VR-based human datasets} \cite{hoque2025egodex}, which utilize on-device SLAM and calibrated cameras to capture hand and wrist pose. VR-based setups exhibit fewer scene constraints and allow more flexible data collection across different environments.
\textbf{(3) Teleoperated robot data} \cite{fourier2025actionnet, qiu2025humanoid}, which involve human operators remotely controlling dexterous robotic hands to execute manipulation tasks. 
\textbf{(4) Self-collected enhanced motion dataset}, as described in \Cref{Wearable System for Enhanced Human Data}, we collected 10K high-fidelity human hand motion trajectories. This enhanced dataset brings three main benefits: (1) it is robust to occlusions and visual ambiguities; (2) it introduces appearance variation via wearable gloves; (3) it provides rich semantic annotations. As shown in \Cref{analysis tokens} (a), the self-collected dataset covers diverse task categories, including single-object and multi-object pick-and-place, assembly, tool use, and articulated manipulation. 57.2\% of trajectories are single-arm and 42.8\% are bimanual.
The distribution is balanced, with higher proportions of tool use (25\%) and precise manipulation (19\%), reflecting an emphasis on complex dexterous skills.

The composition of EgoAtlas dataset is summarized in \Cref{tab:Statistics of EgoAtlas}, which includes 8 sources with detailed annotations of hand motions. In total, EgoAtlas contains 343K trajectories and 89.72M image–action pairs. Weighted sampling is applied to maintain a balanced source distribution.

\section{Motion-aware  dynamics tokens}
\label{Appendix: Motion-aware  dynamics tokens}

To empower the Vision-Language Model (VLM) with enhanced motion comprehension during action execution, we introduce a \textbf{Motion-aware Dynamics Tokens} framework as a supervisory mechanism. This framework is structured to decompose dynamic information into two complementary components: \textbf{Vision Dynamics} and \textbf{Motion Dynamics}.

\noindent
The \textbf{Vision Dynamics} component is designed to learn a compact, discrete representation of the visual change between consecutive frames. This process begins with an encoder $\text{Enc}_V(\cdot)$ based on a self-attention mechanism. The encoder's input is a concatenation of three elements: the visual features $I_t, I_{t+k}$ from the image sequence, the corresponding motion features $P_{t, t+1, ..., t+k}$, and a set of randomly initialized, learnable parameters ${\tilde{D}_{vis}}$:
\[
\hat{D}_{vis} = \text{Enc}_V(I_t, I_{t+k}, P_{t, t+1, ..., t+k}, \tilde{D}_{vis})
\]
These parameters are then discretized using a VQ-VAE objective, resulting in a final representation composed of $V=4$ tokens selected from a candidate codebook of size $|C_v|=16$
\[
D_{vis} = \mathbf{VQ}(\hat{D}_{vis})
\]
This architecture allows the visual information to be deeply infused with motion characteristics. Subsequently, a decoder $\text{Dec}_V(\cdot)$ takes the first frame $I_t$ along with the refined parameters $D_{vis}$ and is tasked with reconstructing the final frame $I_{t+k}$:
\[
\hat{I}_{t+k} = \text{Dec}_V(I_t, D_{vis})
\]
This auto-encoding objective forces $D_{vis}$ to encapsulate the essential visual transformation between the two frames. 

\noindent
Conversely, the \textbf{Motion Dynamics} component aims to capture more complex and granular motion patterns directly from raw action data. We employ a two-layer residual quantization (RQ) architecture for this purpose. To preserve the most authentic motion characteristics, the original action information $M_{t, t+1, ..., t+k}$ is first processed through a Pose Network $\text{PoseNet}(\cdot)$ and then fed directly into the RQ network $\text{RQ}(\cdot)$:
\[
P_{t, t+1, ..., t+k} = \text{PoseNet}(M_{t, t+1, ..., t+k})
\]
\[
D_{mot} = \mathbf{RQ}(P_{t, t+1, ..., t+k})
\]
This architecture selects a total of $R=40$ tokens from a larger, shared codebook of size $|C_m|=512$, enabling a hierarchical and fine-grained representation of motion.
The motion dynamics tokens $D_{mot}$ are then decoded using a Temporal Convolutional Network (TCN) to reconstruct the original motion sequence, serving as the supervision signal to ensure the quantized representation preserves essential temporal dynamics.
\[
\hat{M}_{t, t+1, ..., t+k}  = \text{TCN}(D_{mot})
\]
\noindent
A key design principle is uniformity: all codebooks maintain a unified feature dimension $d=128$. This ensures consistent representation for downstream processing while simultaneously enabling each component to effectively capture the hierarchical spatiotemporal patterns crucial for advanced action-aware modeling.

\section{CoT-Style Reasoning and Acting}
METIS operates in a closed-loop manner with distinct input–output flows depending on the active mode (i.e., Reasoning Mode and Acting Mode).

\begin{itemize}
    \item \textbf{Reasoning Mode:} METIS takes as input: (1) the task language instruction, (2) the current egocentric observation, (3) the latest reasoning. To explicitly signal the transition into reasoning mode, the model first emits a dedicated special token, denoted as $[BOR]$ (\emph{Beginning of Reasoning}). Following this, METIS autoregressively generates a subtask-level description (e.g., use the right hand to pour the juice from the tall blue mug into the bowl.) The generated reasoning tokens do not directly interact with the environment, they are appended to the model's context and maintained as the latest reasoning.
    \item \textbf{Acting Mode:} METIS receives (1) the task language instruction, (2) the current egocentric observation, (3) the latest reasoning, (4) the current unified proprioceptive state. Once triggered by the $[BOD]$ (\emph{Beginning of Dynamics}), VLM autoregressively generates a sequence of motion-aware dynamics tokens, which consist of vision dynamics and motion dynamics. The generated dynamics tokens are subsequently interpreted by an action expert, which maps them into low-level executable control commands.
\end{itemize}

Reasoning Mode and acting mode are realized within a single unified autoregressive framework, where a \emph{token-gated} mechanism determines the active generation stream (i.e., reasoning tokens or acting tokens). This design enables METIS to seamlessly interleave planning and control within a single model, while preserving a consistent input representation and a unified training objective.

\section{Task Details}
\label{Appendix: task details}

\begin{figure*}[h!]
    \centering
    \includegraphics[width=\linewidth]{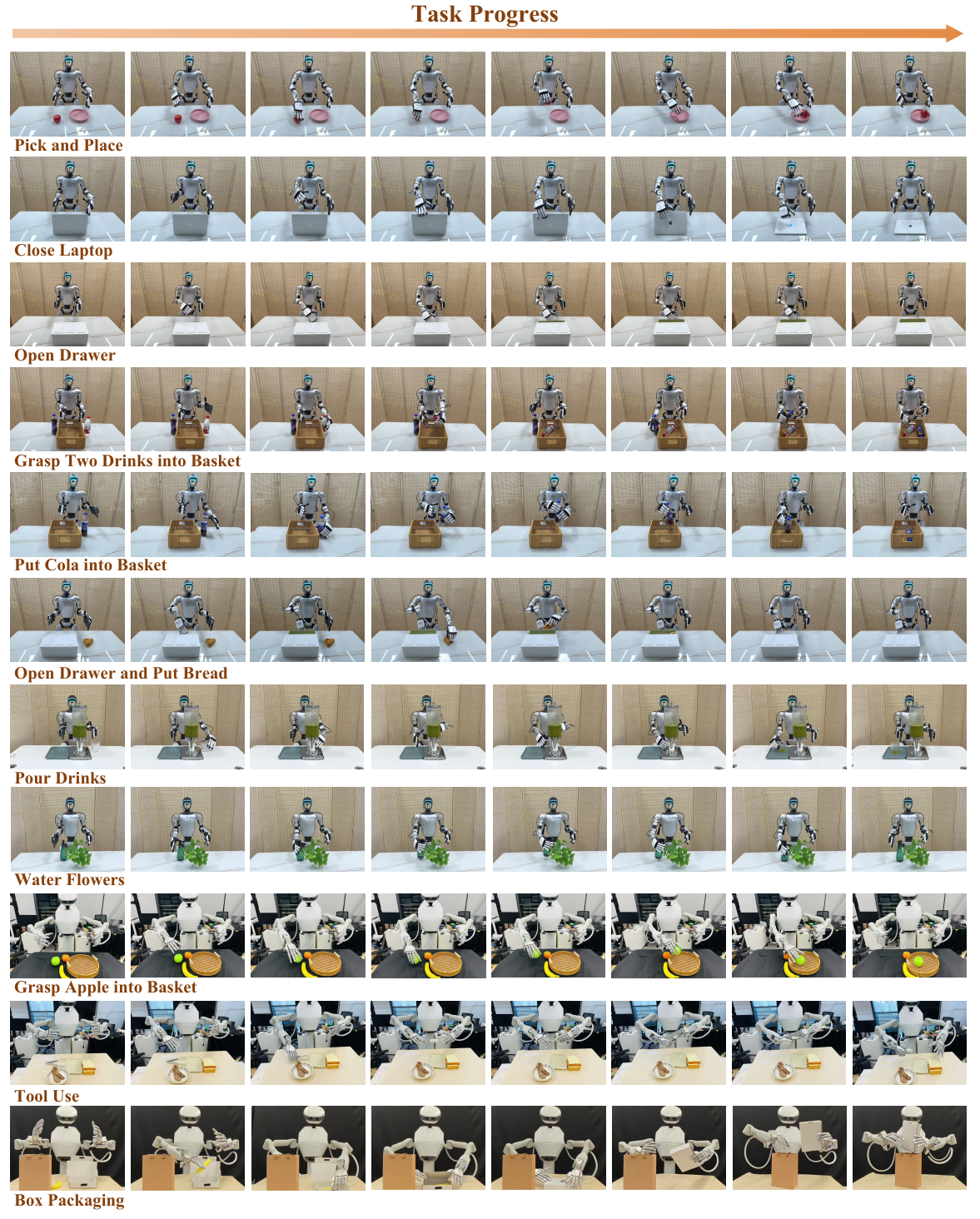}
    \caption{\textbf{Visualization of task progress across embodiments.}}
    \label{fig:task progress}
\end{figure*}

\noindent\textbf{Pick and Place.}
In this task, the robot performs a complete pick-and-place operation involving a common household object—an apple. The robot first identifies the position of the apple on the tabletop through egocentric visual observation and moves its hand toward the target. It then adjusts its wrist orientation and finger configuration to achieve a stable grasp. After successfully lifting the apple, the robot transports it to the target area and gently places it into a plate. This task assesses the model’s capability to perceive and manipulate objects accurately, as well as its ability to coordinate visual perception and fine-grained hand control in a sequential manipulation process. Success is achieved if the apple is successfully placed into the plate.

\noindent\textbf{Close Laptop.} 
In this task, the robot is required to close a partially open laptop. The robot places four fingers along the back edge of the laptop cover and applies a downward motion by rotating the wrist, allowing the lid to close smoothly. This task evaluates the model’s ability to control contact-rich interactions and perform coordinated multi-finger and wrist motions for precise object manipulation. Success is achieved if the laptop is fully closed.

\noindent\textbf{Open Drawer.} In this task, the robot is required to open a partially closed drawer. The robot first bends its four fingers and positions the hand so that the fingertips can reach into the handle of the drawer. It then moves the wrist backward while maintaining a stable grip on the handle, gradually pulling the drawer open. This task requires precise hand movements guided by visual feedback and involves contact-rich manipulation that demands fine coordination between finger articulation and wrist motion. Success is achieved if the drawer is pulled out to a fully open position.

\noindent\textbf{Grasp Two Drinks into Basket.} 
In this bimanual manipulation task, the robot is required to sequentially grasp and place two bottled drinks into a basket. The robot first raises its left hand, reaches toward the bottle, and performs a stable grasp before lifting it and placing it into the basket. It then raises the right hand to repeat the same motion for the second bottle, ensuring symmetric and coordinated control between both hands. After both bottles are successfully placed, the robot closes both hands into fists and pushes the basket forward to complete the task. This task evaluates the model’s ability to perform coordinated bimanual manipulation and to plan sequential actions that involve precise grasping and contact-rich object placement. Success is achieved if both bottles are correctly placed into the basket and the basket is pushed forward. 

\noindent\textbf{Put Cola into Basket.} 
This task requires precise bimanual cooperation to manipulate and place a bottle of cola into a basket. The robot first moves its left hand toward the cola bottle and grasps it from the lower side, lifting it to chest height. The right hand then approaches slowly and grasps the upper part of the bottle, ensuring a stable handover between both hands. After securing the bottle, the robot releases the left hand while the right hand moves toward the basket and places the bottle inside. This task evaluates the model’s ability to perform coordinated bimanual-hand manipulation, including object handover, grasp stability, and visually guided placement. Success is achieved if the cola bottle is stably placed inside the basket.

\noindent\textbf{Open drawer and put bread.}
This long-horizon task requires the robot to perform a sequence of coordinated actions involving both hands. The robot first bends the fingers of its right hand and moves the wrist forward so that the fingertips can reach into the drawer handle. It then pulls the drawer open smoothly through a controlled backward motion. Next, the robot moves its left hand to grasp a piece of bread randomly placed on the tabletop, lifts it, and positions it above the open drawer. The bread is then released into the drawer, after which the robot uses its right hand to close the drawer. This task challenges the model’s ability to perform long-horizon visuomotor control, requiring precise spatial reasoning, bimanual coordination, and contact-rich manipulation across multiple sub-actions. Success is achieved if the bread is successfully placed inside the drawer and the drawer is fully closed.

\noindent\textbf{Pour Drinks.}
In this task, the robot first grasps a cup from one side using its left hand and positions it beneath the beverage dispenser’s outlet. The robot then actuates the dispenser switch with its left hand to pour the drink into the cup. The robot then uses its right hand to grasp the filled cup and place it onto the designated area on the right-side table. This task involves tool use, manipulation of transparent objects, and coordinated bimanual actions, placing high demands on precise control and fine-grained manipulation. Success is achieved when the filled cup is stably placed at the specified target location.

\noindent\textbf{Water Flowers.} In this task, the robot is required to perform fine-grained tool use to water flowers. The robot first grasps a watering sprayer from the table using a functionally appropriate grip, with the thumb supporting the handle while the index and middle fingers are positioned on the trigger. It then aligns the sprayer toward the flower pot and actuates the trigger by pressing the index and middle fingers, causing water to be sprayed onto the plant. This task evaluates the model's ability to execute precise finger-level control, functional grasping , and coordinated tool use. Success is achieved when water is correctly dispensed onto the plant through a stable and functionally correct grasp of the watering tool.

\noindent\textbf{Grasp Apple into Basket.}
In this task, the robot is required to accurately identify the position of an apple among three objects (orange, apple, and banana) present simultaneously in its field of view, then precisely grasp the apple using four fingers. After successfully securing the apple, the robot transports it above a basket and gently places it inside. This task evaluates the model's capability for object perception and manipulation in cluttered environments, as well as its ability to execute targeted grasping and placement operations amid visual distractions. Success is achieved when the apple is securely deposited into the basket.

\noindent\textbf{Tool Use.}
In this task, the robot is required to reposition a bread clip into an optimal grasping posture for functional use. The robot first precisely pinches the bread clip with its right hand, then transfers the tool's end to its left hand for stabilization. Finally, the right hand regrasps the bread clip at an operational angle suitable for precise manipulation. This task evaluates the model's capability for bimanual coordination and its ability to dynamically adjust grasping strategies during tool transfer. Success is achieved when the tool is securely held in a functionally optimal posture through seamless handover between both hands.

\noindent\textbf{Box Packaging.} 
In this task, the robot first grasps a toy banana and places it into a cardboard box. It then folds and assembles the box structure through a series of precise, coordinated bimanual actions. During the process, the robot must handle deformable and semi-rigid materials, continuously adjusting contact points and grasping strategies to maintain structural stability. After the box is fully formed, the robot accurately places the packaged box into a paper bag. This task evaluates the model’s capability for long-term reasoning, multi-stage task decomposition, and dexterous manipulation with high-DoF hands. Success is achieved when the box is fully assembled and correctly packaged through a coherent sequence of grasps, folds, and hand transitions.

More visualization of dexterous manipulation tasks can be found in \Cref{fig:task progress}.

\section{Policy Implementation Details}
\subsection{Pretraining Details}
We use the EgoAtlas dataset to pretrain METIS, which integrates eight heterogeneous data sources covering both human and robot egocentric manipulation trajectories. To ensuring training efficiency and maintain a balanced distribution across data domains, we apply sampling to construct the final training mixture. The detailed dataset composition and mixture weights are provided in \cref{tab:Training Dataset Mixture}.

\begin{table}[h]
    \renewcommand{\arraystretch}{1.15}
    \centering
    \footnotesize 
    \caption{The dataset name and sampling weights used in EgoAtlas pretraining dataset.}
    \label{tab:Training Dataset Mixture}
    \begin{tabular}{c c}
        \toprule
        \multicolumn{2}{c}{\bf Training Dataset Mixture} \\
        \midrule
        H2O \cite{kwon2021h2o} &  0.8\%   \\
        OAKINK \cite{yang2022oakink} & 1.9\%  \\
        PH2D \cite{qiu2025humanoid} & 5.4\%  \\
        ARCTIC \cite{fan2023arctic} & 2.8\%  \\
        EgoDex \cite{hoque2025egodex} & 40.3\%  \\
        Holoassist \cite{wang2023holoassist} & 10.6\%  \\
        ActionNet \cite{fourier2025actionnet} & 13.4\%  \\
        \textbf{Our Enhanced Data} & 25.4\%  \\
        \bottomrule
    \end{tabular}
\end{table}

During pretraining, we jointly optimize all model parameters, including the vision encoder, the LLM backbone, and the action decoder. We use a global batch size of 768 (32 per device) and use distributed training under a fully sharded data-parallel (FSDP) setup. Optimization is performed with AdamW using a learning rate of $2e^{-5}$, no weight decay, and gradient clipping at a max-norm of 1.0. The action loss weight $\lambda$ is set to 1. We train on a cluster of 24 NVIDIA H100 GPUs. Empirically, we find that 60k training steps are sufficient to achieve strong performance, which requires approximately 72 hours.

\subsection{Posttraining Details}
For post-training, we use an 8-GPU setup with a per-device batch size of 4. We apply LoRA updates with a rank of 32 to both the LLM backbone and the vision encoder, while performing full-parameter fine-tuning on the action decoder. Only the LoRA-adapted weights and the action decoder parameters are included in the trainable set. We optimize the model using AdamW with a learning rate of $3.5e^{-4}$ and a weight decay of $1e^{-3}$. The learning rate schedule follows a StepLR policy, where the learning rate is decayed by a factor of 0.1 after 80\% of the total training steps. We train the model for 80000 steps in total. To improve training efficiency, we adopt mixed-precision training: both the LLM backbone and the vision encoder are trained in bfloat16 precision, whereas the action decoder is kept in float32 for numerical stability. This configuration provides a favorable trade-off between speed and accuracy during post-training.

For the robot observation–action interface, we set the proprioceptive history to 1 and use an action chunk length of 32. Real-world data are collected at 30 Hz. The recorded images are first center-cropped to a resolution of $480\times 640$, and subsequently resized to $224\times224$ before being fed into the visual encoder.

\subsection{Baseline Settings}

\noindent\textbf{For OpenVLA-OFT \cite{kim2025fine}}, all experiments are conducted on 4 NVIDIA H100 GPUs. We use the full joint configuration of the dual-arm and dual-hand system as the policy action space (2*7 DoF for arm joints and 2*6 DoF for hand joints). We apply LoRA-based fine-tuning to the VLA and adopt an L1 regression–style action head for continuous action prediction. The action chunk length is set to 30. We use the following hyperparameters for training OpenVLA-OFT.

\begin{table}[h]
    \renewcommand{\arraystretch}{1.15}
    \centering
    \footnotesize 
    \caption{Hyperparameters for OpenVLA-OFT.}
    \label{tab:hyperparameters for OpenVLA-OFT}
    \begin{tabular}{c c}
        \toprule
        \bf Name & \bf Value \\
        \midrule
        Steps &  40000   \\
        Batch Size & 4  \\
        Learning Rate & $5e^{-4}$  \\
        Action Chunks & 30  \\
        LoRA Rank & 32  \\
        Action Head & $l1_regression$  \\
        \bottomrule
    \end{tabular}
\end{table}

\noindent\textbf{For $\pi_{0.5}$ \cite{intelligence2025pi_}}, post-training uses 1 NVIDIA H100 GPU across all tasks. We use wrist pose for arms and joints for dexterous hands (2*7 DoF for wrist poses and 2*6 DoF for hand joints). Action chunk size is set to 10. The hyperparameter configuration is as follows.

\begin{table}[h]
    \renewcommand{\arraystretch}{1.15}
    \centering
    \footnotesize 
    \caption{Hyperparameters for $\pi_{0.5}$.}
    \label{tab:hyperparameters for pi0.5}
    \begin{tabular}{c c}
        \toprule
        \bf Name & \bf Value \\
        \midrule
        Steps &  30000   \\
        Batch Size & 32  \\
        Learning Rate & $5e^{-5}$  \\
        Action Chunks & 10  \\
        Ema decay & 0.999 \\
        Warm-up steps & 10000 \\
        \bottomrule
    \end{tabular}
\end{table}

\noindent\textbf{For GR00T N1.5 \cite{bjorck2025gr00t}}, All experiments are conducted on 4 NVIDIA H100 GPUs. We use joint angles as the policy action space, which contains two 7-DoF arm joints and two 6-DoF hand joints. Fine-tuning is set to freeze the language model backbone and vision tower, only fine-tuning the projector and diffusion model. The action chunk length is set to 16, and some hyperparameters are set as follows.

\begin{table}[h]
    \renewcommand{\arraystretch}{1.15}
    \centering
    \footnotesize 
    \caption{Hyperparameters for GR00T N1.5.}
    \label{tab:hyperparameters for OpenVLA-OFT}
    \begin{tabular}{c c}
        \toprule
        \bf Name & \bf Value \\
        \midrule
        Steps &  10000   \\
        Batch Size & 32  \\
        Learning Rate & $1e^{-4}$  \\
        Weight Decay & $1e^{-5}$ \\
        Warmup Ratio & $0.05$ \\
        Action Chunks & 16  \\
        \bottomrule
    \end{tabular}
\end{table}

\noindent\textbf{For ACT \cite{zhao2023learning}}, we choose to follow the lerobot \cite{cadene2024lerobot} framework. All training is conducted on a single NVIDIA H100 GPU. At this point, we unify the action space with METIS, using the 9D pose of the wrist and the 3D XYZ pose of the fingers as the action space. We use ResNet18 as the vision backbone for training, with the remaining transformer layers trained from scratch. The action chunk length is set to 100, and the hyperparameters are as follows.

\begin{table}[h]
    \renewcommand{\arraystretch}{1.15}
    \centering
    \footnotesize 
    \caption{Hyperparameters for ACT.}
    \label{tab:hyperparameters for OpenVLA-OFT}
    \begin{tabular}{c c}
        \toprule
        \bf Name & \bf Value \\
        \midrule
        Steps &  600000   \\
        Batch Size & 8  \\
        Learning Rate & $1e^{-5}$  \\
        Weight Decay & $1e^{-4}$ \\
        Warmup Ratio & $0.05$ \\
        Action Chunks & 100  \\
        Vision Backbone & ResNet-18 \\
        \bottomrule
    \end{tabular}
\end{table}

For ACT, we adopt the same action space representation as METIS to enable a fair comparison. Instead, existing pretrained VLA models, including OpenVLA-OFT, $\pi_{0.5}$, and GR00T N1.5, are originally trained in joint-based action spaces. To avoid disrupting their learned action hierarchies and pretrained representations, we retain their native joint-level action formulations. All baselines are trained on the same datasets and evaluated under identical experimental conditions.
 
\section{Cross-Embodiment Settings}
In the cross-embodiment experiments, we employ the Sharpa Beta embodiment equipped with a pair of 22-DoF SharpaWave Dexterous Hands. During the data collection phase, we capture first-person view images at 30 Hz along with joint state-action pairs at 60 Hz. The data processing pipeline begins with temporal alignment between the 30 Hz images and 60 Hz joint angles, synchronizing both modalities to 30 Hz. For the joint angle information, we perform forward kinematics (FK) calculations and coordinate frame transformations: the fingertip poses (3-dimensional) are transformed into the wrist coordinate frame, while the wrist poses (9-dimensional) are converted to the camera coordinate frame, ensuring a unified action space. During inference, the model's output fingertip and wrist poses undergo inverse kinematics (IK) computations to convert them back to joint angles for robot control.

\section{Quantitative Results and Visualization}
\label{Quantitative Results and Visualization}
\begin{figure}[H]
\caption{\bf (a) Task categories and distribution of the collected dataset. 
(b) Quantitative analysis of code usage with dynamics tokens. 
(c) Quantitative analysis of motion reconstruction with dynamics tokens.}
\label{analysis tokens}
\centering
\includegraphics[width=1\textwidth]{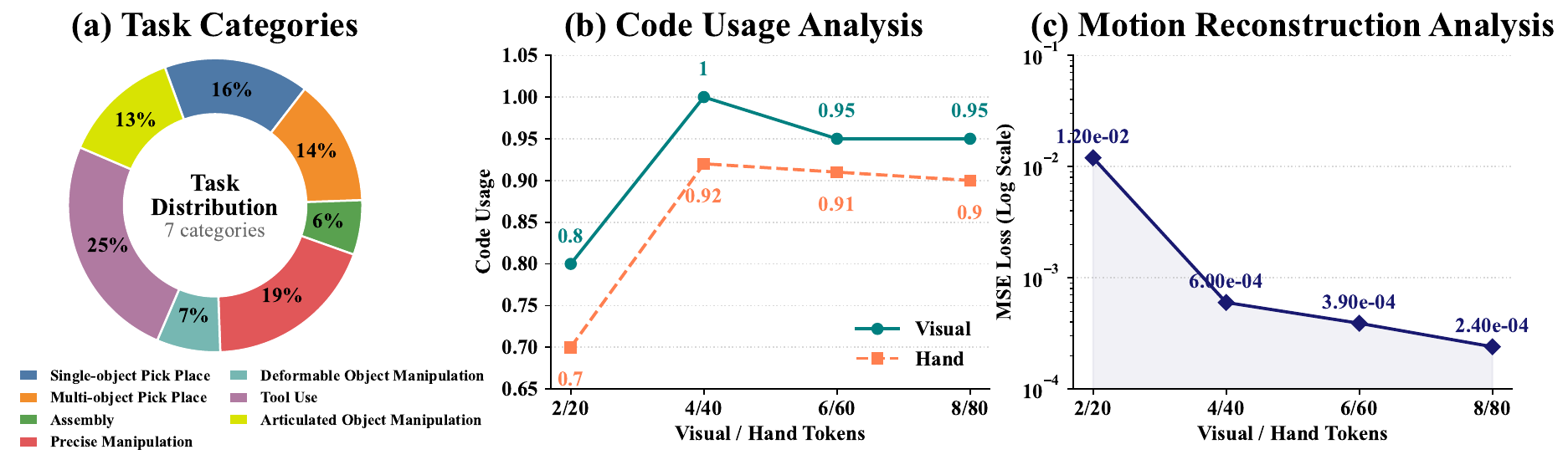}
\label{fig:example}
\end{figure}

We analyze the effect of motion-aware tokens on code usage and motion reconstruction under different token budgets as shown in \Cref{analysis tokens} (b) and (c). Increasing the number of dynamics tokens improves reconstruction fidelity but also incurs higher inference cost, revealing a clear accuracy–efficiency trade-off. Our 4+40 setting strikes a favorable balance, achieving strong reconstruction accuracy while maintaining efficient code utilization and low latency. Further increasing the token budget yields only marginal gains in motion reconstruction quality.

 \vspace{-0.3cm}

\begin{figure}[H]
\centering
\caption{\textbf{Visualization of Codewords.}}
\label{Visualization of Codewords}
\includegraphics[width=1\textwidth]{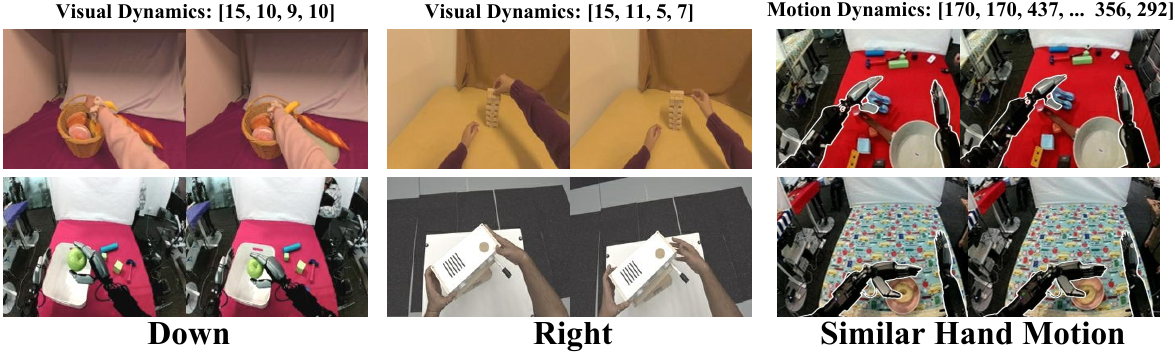}
\label{fig:dynamics_visualization}
\end{figure}

We also visualize learned codewords in \Cref{Visualization of Codewords}.  Notably, across different embodiments and scene contexts, consistent motion trends (e.g., downward or rightward movements) and similar hand pose are quantized into the same codewords, despite variations in visual appearance. This suggests that motion-aware tokens capture underlying motion semantics rather than surface-level configurations, enabling invariance to embodiment and environmental changes. 
Such structured tokenization indicates that the learned representation forms a set of reusable motion primitives, which facilitates efficient encoding and generalization across diverse manipulation scenarios.

\section{Failure Case Analysis}
Through extensive real-world experiments, we identify two primary categories of failure modes that can adversely affect the performance of METIS. The first category of failures arises from the inherent limitations of a \textbf{single egocentric viewpoint}. In tasks that require large-range manipulations or precise interactions with small objects, the restricted field of view often causes target objects to move outside the camera’s visible region or become partially occluded. Such visibility degradation can lead to inaccurate perception and, consequently, suboptimal or failed action execution. The second failure mode stems from \textbf{Inverse Kinematics error}. When controlling high DoF dexterous hands, deriving joint configurations solely from fingertip positions via IK can yield singular or ill-conditioned solutions. Such instability may manifest as unintended finger oscillations or jitter during execution, which in turn degrades manipulation precision and can lead to task failure. Introducing additional kinematic constraints into the IK solver or adopting a learning-based IK predictor may be potential mitigation strategies.

\section{Limitations}
\label{Limitations}
Despite the exceptional performance demonstrated by METIS, several limitations remain. First, our model relies solely on egocentric observations, which may restrict its ability to perceive complete object geometry and fine interaction details. This limitation could be mitigated by incorporating additional wrist-mounted or external cameras. Second, the pretraining process currently excludes large-scale third-person data available online. Extending pretraining to broader multi-view manipulation datasets represents a promising direction for future work.



\end{document}